\documentclass{article}

\usepackage[numbers]{natbib}
\usepackage{PRIMEarxiv}

\usepackage[utf8]{inputenc} 
\usepackage[T1]{fontenc}    
\usepackage{hyperref}       
\usepackage{url}            
\usepackage{booktabs}       
\usepackage{amsfonts}       
\usepackage{nicefrac}       
\usepackage{microtype}      
\usepackage{lipsum}
\usepackage{fancyhdr}       
\usepackage{graphicx}       
\graphicspath{{media/}}     
\usepackage{amsmath}
\usepackage{amssymb}
\usepackage{breqn}
\usepackage{paralist}
\usepackage[ruled,vlined]{algorithm2e}
\usepackage{xcolor}
\usepackage{subfig}
\usepackage{comment}
\usepackage{appendix}
\usepackage{multirow}
\usepackage{siunitx}

\newcommand{\frcnn}{Faster R-CNN\xspace}

\newcommand{\fcos}{FCOS\xspace}
\newcommand{\dgfcos}{DGFCOS\xspace}
\newcommand{\yolo}{YOLOv3\xspace}
\newcommand{\dgyolo}{DGYOLO\xspace}
\newcommand{\dgfrcnn}{DGFR-CNN\xspace}
\newcommand{\etal}{\textit{et al.}}

\definecolor{d1}{RGB}{255,0,0}
\definecolor{d2}{RGB}{0,255,0}
\definecolor{d3}{RGB}{0,0,255}
\definecolor{d4}{RGB}{0,255,255}
\definecolor{d5}{RGB}{0,150,0}
\definecolor{d6}{RGB}{150,0,0}
\definecolor{d7}{RGB}{150,150,0}

\definecolor{c}{RGB}{0,128,0}

\newcommand\crule[3][black]{\textcolor{#1}{\rule{#2}{#3}}}

\newcommand{\person}[1]{\color[RGB]{255,0,0}Red}
\newcommand{\Rider}[1]{\color[RGB]{0,255,0}Green}
\newcommand{\Car}[1]{\color[RGB]{0,0,255}Blue}
\newcommand{\Truck}[1]{\color[RGB]{0,255,255}Cyan}
\newcommand{\Bus}[1]{\color[RGB]{255,255,0}Yellow}
\newcommand{\Train}[1]{\color[RGB]{255,0,255}Violet}
\newcommand{\Motorcycle}[1]{\color[RGB]{150,0,0}Dark Red}
\newcommand{\Bicycle}[1]{\color[RGB]{0,150,0}Dark Green}

\usepackage[capitalize]{cleveref}
\crefname{section}{Sec.}{Secs.}
\Crefname{section}{Section}{Sections}
\Crefname{table}{Table}{Tables}
\crefname{table}{Tab.}{Tabs.}

\title{Domain Generalisation for Object Detection \\ under Covariate and Concept Shift
}

\author{
  Karthik Seemakurthy  \\
  Applied AGI Limited,
  London, UK.\\
  \texttt{karthik.seemakurthy@appliedagi.com} \\
   \And
  Erchan Aptoula \\
  Faculty of Engineering and Natural Sciences (VPALab) \\
  Sabanci University,   Istanbul, Türkiye \\
  \texttt{erchan.aptoula@sabanciuniv.edu} \\
  \And
  Charles Fox, Petra Bosilj \\
  School of Computer Science
  \\
  University of Lincoln, UK\\
  \texttt{\{chfox, pbosilj\}@lincoln.ac.uk} \\
}

\begin{document}
\maketitle

\begin{abstract}
Domain generalisation aims to promote the learning of domain-invariant features while suppressing domain-specific features, so that a model can generalise better to previously unseen target domains. An approach to domain generalisation for object detection is proposed, the first such approach applicable to any object detection architecture. Based on a rigorous mathematical analysis, we extend approaches based on feature alignment with a novel component for performing class conditional alignment at the instance level, in addition to aligning the marginal feature distributions across domains at the image level. This allows us to fully address both components of domain shift, i.e. covariate and concept shift, and learn a domain agnostic feature representation.
Extensive evaluation with both one-stage (\fcos, \yolo) and two-stage (\frcnn) detectors, on a newly proposed benchmark comprising several different datasets for autonomous driving applications (Cityscapes, BDD10K, ACDC, IDD) as well as the GWHD dataset for precision agriculture, shows consistent improvements to the generalisation and localisation performance over baselines and state-of-the-art.
\end{abstract}



\section{Introduction}

\begin{figure}[ht]
\begin{center}
\subfloat[\emph{ACDC}]{\label{fig.acdc}\includegraphics[width=0.23\textwidth]{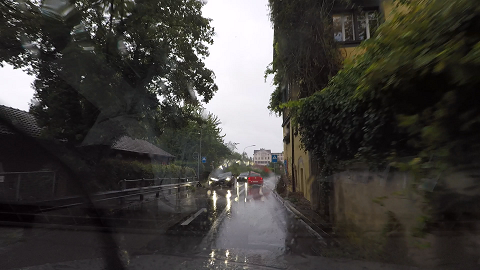}}\,
\subfloat[\emph{BDD10k}]{\label{fig.bdd}\includegraphics[width=0.23\textwidth]{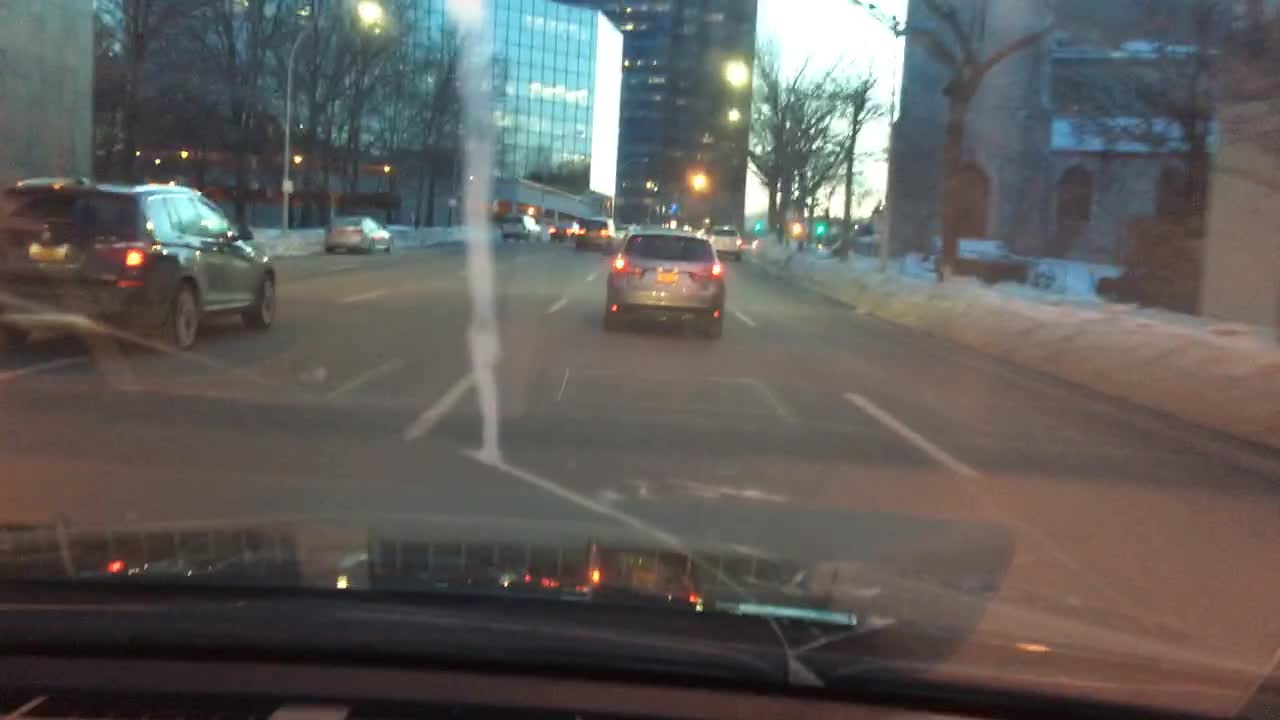}}\,
\subfloat[\emph{Cityscapes}]{\label{fig.cityscapes}\includegraphics[width=0.23\textwidth]{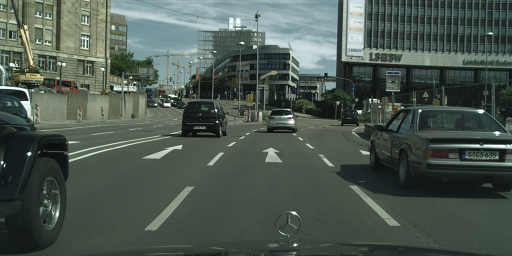}}\,
\subfloat[\emph{IDD}]{\label{fig.idd}\includegraphics[width=0.23\textwidth]{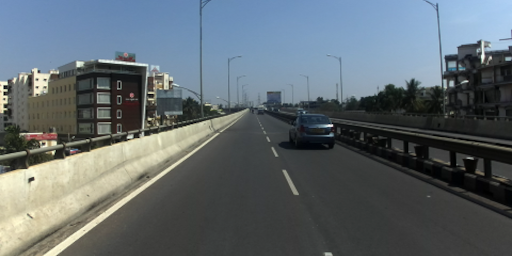}}\,
\subfloat[\emph{GWHD} (France)]{\label{fig.gwhd0}\includegraphics[width=0.23\textwidth]{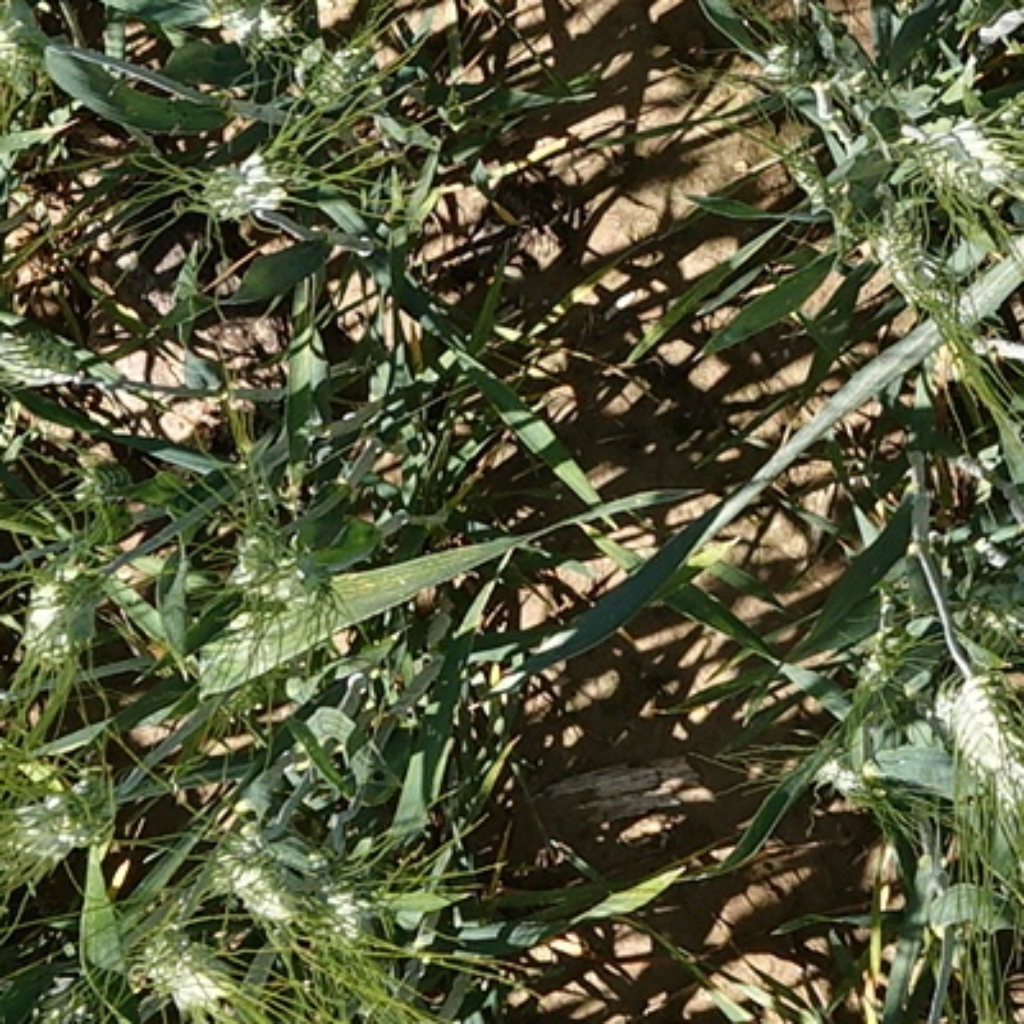}}\,
\subfloat[\emph{GWHD} (Belgium)]{\label{fig.gwhd1}\includegraphics[width=0.23\textwidth]{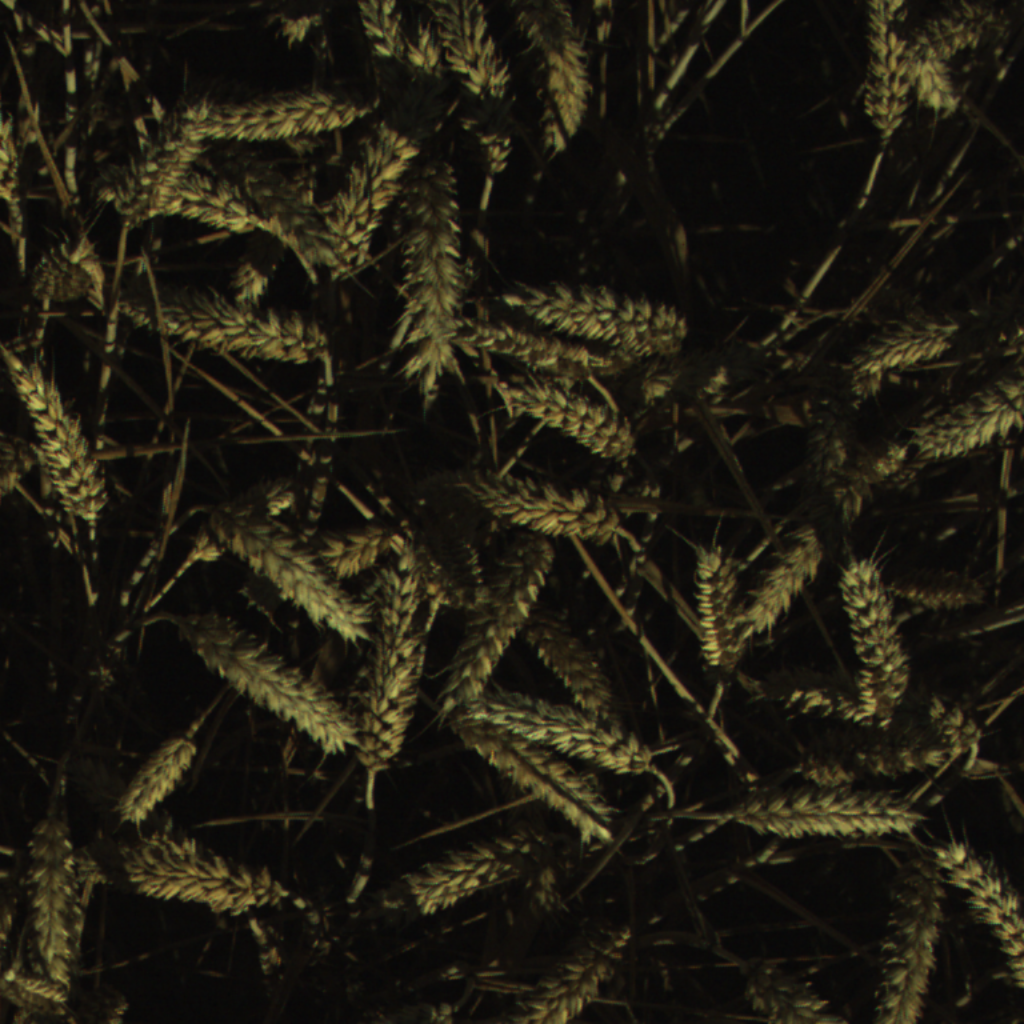}}\,
\subfloat[\emph{GWHD} (Norway)]{\label{fig.gwhd2}\includegraphics[width=0.23\textwidth]{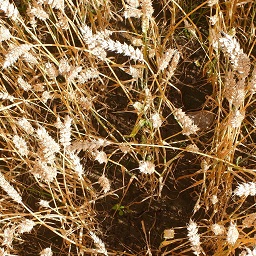}}\,
\subfloat[\emph{GWHD} (Switzerland)]{\label{fig.gwhd3}\includegraphics[width=0.23\textwidth]{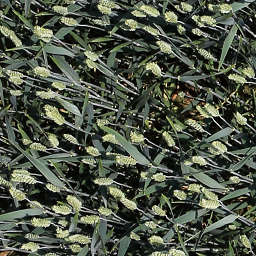}}
\end{center}
\caption{Samples from various datasets used in our experiments.}
\label{fig:domain_shift}
\end{figure}

Object detection (OD) is the task of identifying and localising all instances of a certain type of object in images. Benchmark performances have increased significantly using deep learning approaches    \cite{jocher2020yolov5,tan2020efficientdet,liu2016ssd,ren2015faster,lin2017feature,jifeng2016rfcn, everingham2010pascal,lin2014microsoft}. However, deploying models in autonomous systems continues to be challenging, as factors such as viewpoint, background, weather, and capture devices compound with the variations in object appearance. The resulting distribution discrepancy between training and testing data is called {\em domain shift} and degrades model performance at deployment \cite{recht2019imagenet,hendrycks2019benchmarking}. Fig.~\ref{fig:domain_shift} shows examples illustrating the domain shift that can be encountered in autonomous farming and driving contexts. One can observe significant levels of variations across the appearance of wheat and vehicles in terms of size, shape, illumination, and their positioning in the scene, made more challenging by changing weather conditions.

\begin{figure}
    \begin{center}
    \subfloat[Covariate shift]{\label{subfig:covariate}\includegraphics[width=0.45\textwidth]{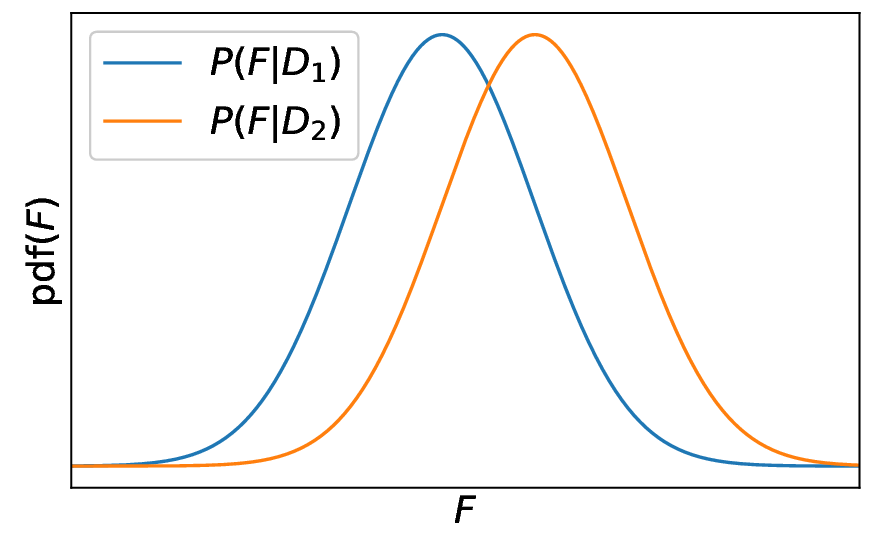}}\,
    \subfloat[Concept shift]{\label{subfig:concept}\includegraphics[width=0.45\textwidth]{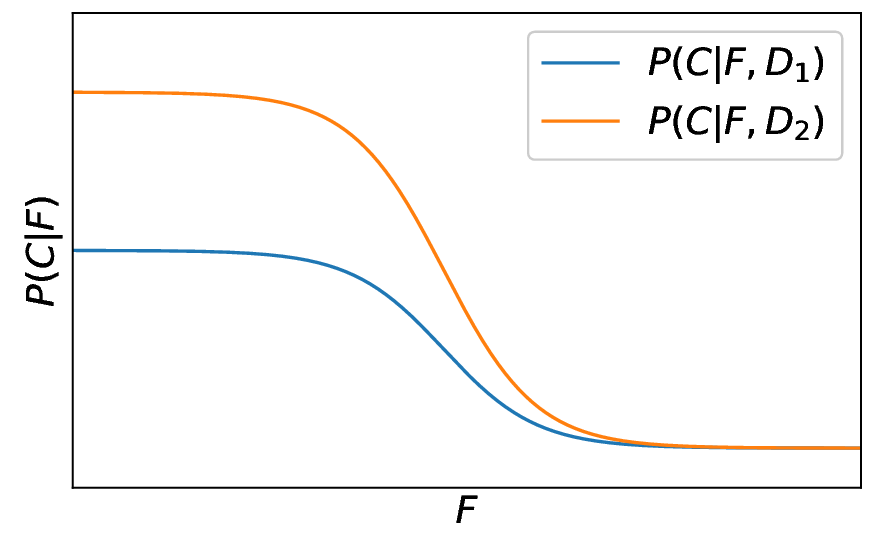}}\\
    \end{center}
    \caption{Different types of domain shift which can describe the relationship between two domains $D_1$ and $D_2$, on a simplified example of using a single feature $F$ to predict the existence of a class $C$. Covariate shift, shown in \protect\subref{subfig:covariate}, describes the situation when the feature (input data) distribution changes between domain. Concept shift, shown in \protect\subref{subfig:concept} describes the situation where the relationship between the class $C$ and the features $F$ changes. The overall distribution shift between two domains is typically a combination of covariate and concept shift.}
    \label{fig:types_of_shift}
\end{figure}

Alleviating the impact of domain shift by increasing the amount and diversity of annotated training data is both expensive and time consuming, and still provides no guarantee that all deployment scenarios will be adequately represented.
In practice, training data obtained from the specific deployment (target) domain is often sparse or unavailable. 
Therefore, \emph{domain generalisation} (DG) addresses the domain shift by dividing the training data into several known domains during training to learn how to to suppress domain-specific information. The goal is to directly learn a single, unchanging set of parameters which are able to perform well on previously unseen domains.
However, recent surveys on the topic of DG \cite{zhou2021domain,koh2021wilds} indicate the lack of a systematic analysis of DG for OD. This paper offers a rigorous mathematical treatment of feature alignment approaches in OD while
addressing both components of domain shift, namely covariate and concept shift (Fig.\@~\ref{fig:types_of_shift}, Eq.\@~(\ref{eq:covcon})).
Covariate shift describes the change in the input distribution, while concept shift refers to the change in the relationship between the predictive features and the expected output \cite{moreno2012unifying}. 
Note that there is an alternative model of domain shift, briefly shown later in Eq.\@~(\ref{eq:labman}), decomposing it into label and manifestation shifts, which we do not use in this paper.
We show that aligning the marginal feature distributions across domains at the image level \cite{liu2020towards}, and even the instance level \cite{chen2018domain} only addresses the covariate shift, which is necessary, but not sufficient to promote DG in OD. We further show that addressing concept shift by aligning class conditional distributions at the instance level, as well as ensuring consistency between instance and image level features \cite{seemakurthy2023icra,seemakurthy2023aaai} consistently improves DG for OD. This paper brings together and builds upon the previous work by the authors \cite{seemakurthy2023icra,seemakurthy2023aaai} to show how our proposal to jointly address both the covariate and concept components of domain shift can be applied to any object detection architecture, the first architecture-agnostic DG-OD approach. We include an extensive evaluation of our approach on both one-stage (\yolo, FCOS) and two-stage (\frcnn) object detection architectures, 4 autonomous driving datasets and 1 autonomous farming dataset, improving the performance
of all baseline architectures and outperforming the state-of-the-art approaches \cite{liu2020towards,lin2021domain} on their own benchmarks. Finally, we propose an extended and improved benchmark for DG in OD for autonomous driving, relying on 4 distinct and independent datasets (compared to the one used by \cite{lin2021domain}, which only uses 3 datasets, one of which is synthetically derived from another). Our implementations and models are open-sourced to encourage further research.

\section{Related Work}
\label{sec:related}


\textbf{Object detection.} Classical approaches to OD relied on handcrafted features and formulated OD as a sliding window classification problem \cite{dalal2005histograms,felzenszwalb2009object,viola2001rapid}, combining classification and localisation in a single task. The empirical success of deep neural networks (DNNs) -- which automatically extract high performing features -- \cite{lecun1998gradient,krizhevsky2012imagenet} has resulted in their widespread adoption.  
Most CNN-based approaches to object detection can be categorised as either single-stage \cite{jocher2020yolov5,tan2020efficientdet,liu2016ssd,redmon2018yolov3,redmon2016you,tian2019fcos,carion2020end,He2022CVPR,10204463,10204055} or two-stage \cite{ren2015faster,lin2017feature,jifeng2016rfcn,girshick2014rich,girshick2015fast}, but OD can still be interpreted as a combination of a classification and localisation task. 
{\em Two-stage} approaches developed from the Region-based CNN (R-CNN) family \cite{ren2015faster,girshick2014rich,girshick2015fast} 
are characterised by a network trained to classify region proposals selected from the image in a separate preceding localisation step. {\em Single-stage}  approaches perform localisation and classification simultaneously and have only recently reached the performance of two-stage approaches \cite{jocher2020yolov5}. However, all of these architectures and a number of follow-up works \cite{liu2016ssd,lin2017feature,jifeng2016rfcn} operate under the assumption that the testing data originates from the same domain and distribution as the training data and their performance consequently degrades on out-of-domain (OOD) data. When addressing domain shift, it is important to note that different to classification, where instance and image features are one and the same, in OD image level domain features can have a substantially different distribution to instance level class features. Therefore, our treatment of domain shift for OD addresses shift jointly at the image and instance level.

\textbf{Domain shift.} A number of techniques exist to directly or indirectly address domain shift. Transfer learning (TL) \cite{pan2010survey} is one of the best-known examples, enabled by a high transferability of pre-trained deep features, especially from the lower layers. Architectures pre-trained on a large corpus such as ImageNet \cite{deng2009imagenet} are fine-tuned for a different but related task. Contrary to DG, TL requires annotated data from the target domain, but can decrease both the amount of time and annotations needed by several orders of magnitude compared to training from scratch. The shift comes both from the distribution change between source and target domains, and a disjoint label space. It has been noted that the effectiveness of TL drops with increasing domain shift \cite{zhuang2020comprehensive,bosilj2020transfer,ghazi2017plant,niu2020decade}.
\emph{One-shot} and \emph{few-shot learning} (OSL and FSL) \cite{wang2019panet} are a special case of transfer learning, where the representation of a new class is learned only from a single (or a handful of) labelled image(s) of that class. 
Unlike TL and OSL/FSL approaches which require fully annotated target domain data, \emph{self-supervised learning} (SSL) \cite{jing2020self} approaches leverage unlabelled data to minimise the shift in feature space. This can be realised either sequentially, as a pre-training step followed by a classical supervised TL \cite{noroozi2016unsupervised,hindel2023inod}, or in a semi-supervised multi-task learning setup, where the unlabelled data is used in conjunction with a (typically much smaller quantity of) labelled data \cite{ilteralp2021deep,ullah2023ssmd,li2022learning}. Since the secondary task, relying on unlabelled data, has a reguralising effect on the learned features \cite{yang2016deep}, this paradigm has also been applied for DG \cite{wang2020learning}. \emph{Domain randomisation} \cite{tobin2017domain,yue2019domain} is a complementary approach to DG, which relies on synthetically generated variations of the input data to obtain more generalisable features. \emph{Domain adaptation} (DA) \cite{chen2018domain,saenko2010adapting} the closest topic to DG, as they both aim to address the domain shift encountered in new environments (target domain distribution shift) without the change to the label space. However, unlike DG, where no information about the target data distribution is known, DA relies on sparsely labelled or unlabelled target domain data (or other information providing insight into target data distribution) during training.

\textbf{Domain generalisation in classification.} DG was first studied \cite{blanchard2011generalizing} in the context of medical imaging, while the terminology was introduced later \cite{muandet2013domain}. Earlier studies have explored fixed shallow features \cite{muandet2013domain,khosla2012undoing,xu2014exploiting,fang2013unbiased,ghifary2015domain}, while more recent investigations design architectures to address domain shift \cite{li2017deeper} or learning algorithms to optimise standard architectures \cite{li2018learning,shankar2018generalizing,li2019episodic}. One of the most common approaches to DG (as well as DA) is \emph{feature alignment} \cite{muandet2013domain,ganin2016domain,li2018deep}, where the aim is to minimise the difference between the features representations of the source domains in order to learn a domain-invariant representation. Typical alignment techniques include domain-adversarial learning \cite{li2018deep}, moments \cite{ghifary2016scatter}, contrastive loss \cite{mahajan2021domain}, KL-divergence \cite{li2020domain} and maximum mean discrepancy \cite{li2018domain}. Contrary to feature alignment which attempts to make the whole model domain-invariant, in approaches based on \emph{representation disentanglement} \cite{li2017deeper,ilse2020diva}, the models contain both domain-specific and domain-agnostic parts.
In \emph{meta-learning}, training data is split into meta-train and meta-test sets which exhibit domain shift akin to the one expected at test-time, with the goal of learning how to train using the meta-train set in a way that improves the performance on the meta-test set. MAML \cite{finn2017model}, a meta-learning technique for DA, has been repurposed for DG \cite{li2018learning}, and extended in numerous follow-up works \cite{dou2019domain,du2020metanorm}. Approaches performing advanced \emph{data augmentations} attempt to diversify the training data, e.g.\@~by applying off-the-shelf style transfer models \cite{shi2022gradient} or learning to synthesise new domains \cite{zhou2020learning}. A more detailed review \cite{zhou2021domain} covers different domain generalisation approaches.

\textbf{Domain generalisation beyond classification.} Despite the numerous advances in understanding and addressing domain shift, there are not many works dealing with DG, or domain shift at all, in tasks beyond classification. A DG approach based on meta-learning was proposed for semantic segmentation \cite{li2018learning}, however, different to the typical DG setting, this approach requires additionally processing the target domain data at inference time. DG has also been applied to person re-identification \cite{song2019generalizable} and face presentation \cite{wang2020cross,jia2020single}, however both of these applications have a key characteristic in common with classification, namely the fact that the image-level and object-level features are one and the same.
\citet{chen2018domain} propose a DA approach for minimising the domain shift in OD, a task which distinguishes between image and instance-level features. Their approach is based on feature alignment, and attempts to perform feature alignment consistently between the image and instance-level features, similar to our proposal. However, while it can be applied to DG with minimal modifications, it relies on the assumption that the class distribution and classifier behaviour is consistent across domains. This assumption has been shown to be false, especially when the shift between input image domains is significant \cite{li2018deep,li2018domain, zhao2020domain, scholkopf2012causal, janzing2010causal,hu2020domain}, which has been addressed by this work, using a similar approach to that which \citet{zhao2020domain} applied to classification.
DG for OD has been explicitly addressed only by \citet{liu2020towards} (which was extended by \citet{chen2023achieving}) and \citet{lin2021domain}. In common with the work proposed herein, \citet{liu2020towards} also rely on feature alignment to achieve DG, however, unlike our work, they only address covariate shift by aligning the image-level features, similarly to classification approaches. In addition, they rely on style transfer to generate new domain examples, and implement the feature alignment jointly through the Gradient Reversal Layer (GRL)\cite{ganin2016domain} (like us) and Invariant Risk Minimisation (IRM) \cite{arjovsky2019invariant} (which we do not do). The follow-up work to \citet{liu2020towards}, proposed by \citet{chen2023achieving}, improves the new domain synthesis, and implements feature alignment through contrastive learning, however still only addresses the covariate shift at the image level. Like our own work, the approach for \citet{lin2021domain} is tailored specifically for OD, addressing domain shift both at the image and instance level.
However, it conflates two incompatible domain shift models (Eqs.\@~\ref{eq:covcon} and \ref{eq:labman}) by attempting to address covariate and label shift at the same time.
Finally, while all of these DG for OD approaches are tailored for one specific architecture (\citet{liu2020towards,chen2023achieving} work on \yolo \cite{redmon2018yolov3}, while \citet{lin2021domain} work with \frcnn \cite{ren2015faster}), the work presented in this paper constitutes the first DG for OD approach which can be used with any detector.

\section{Preliminaries}
\label{sec:generalised_object_detection}

In this section the DG task is formally defined, and the mathematical context used by the proposed solution is presented.



Let a \textit{model}
be any function which takes an image as input, and transforms it into a feature vector according to its parameters, and returns a set of bounding boxes and a vector of class probabilities for each bounding box, based only on these features and additional parameters. Let a {\em trainer} be any system which takes images, ground truth bounding boxes containing objects of interest as well as their respective classes as training data, a model, and a loss function quantifying the system's performance in terms of classification accuracy and bounding box prediction quality, all as input; and returns a set of model parameters which minimises the loss function between the model outputs and the training data, called a \emph{trained model}. 

DG assumes that the training data is generated from multiple domains, and labelled with these domains in addition to the bounding boxes and classes. A {\em DG trainer} is a system which takes similar inputs to a trainer with these domain labels added to images, and returns a trained model capable of generalising to unseen target domains without compromising the classification or bounding box prediction performance. 

Let $\mathbf{I} \times \mathbf{C} \times \mathbf{B} \times \mathbf{D}$ be the sample space under observation. 
Here $I \in \mathbf{I}$ is an image, and $B \in \mathbf{B}=\mathbb{R}^{4}$ is a bounding box as a tuple  $(x,y,w,h)$, of its central coordinate, width and height.  $C \in \mathbf{C}$ denotes the detection class. 
$D \in \mathbf{D}$ is the domain to which an image belongs, $|\mathbf{C}|$ is the total number of classes in the training data, $|\mathbf{D}|$ is the total number of domains in the training data, and $|D|$ is the number of images in domain $D$.  
Let $P \left(I, C, B | D\right)$ be the ground truth joint distributions defined on $\mathbf{I} \times \mathbf{C} \times \mathbf{B}$ given domain $D$. The DG trainer's task is to take as input a model and ground truth data from $\mathbf{I} \times \mathbf{C} \times \mathbf{B} \times \mathbf{D}$, and return a trained model capable of domain invariant detection.

 

Let a model be defined by the tuple $(f,c,b,\phi, \gamma, \beta)$ of functions and parameters for a feature vector extractor $F=f(I;\phi)$, a classifier on these features, $c(F; \gamma) \approx P(C|I)$, and a bounding box predictor $b(F; \beta)\approx P(B|I)$, respectively. 
Let $Q(C,B|I)$ be the joint distribution induced by a model given an image. The DG trainer's  task is to optimise the parameters in order to best approximate $Q(C,B|I) \approx P(C,B|I)$.  This equation is independent of domain $D$.   

{\em Notation conventions}: $Q$ is parameterised by -- and so depends on -- all of the    parameters, but we omit them in the notation for brevity unless specified otherwise.  All distributions involving $F$ are also parameterized by $\phi$ which we also omit from notation for the sake of brevity. Expectations, entropies, and other functionals are assumed to be taken over the conditioned, not conditioning variables unless otherwise stated, for example $\langle P(A,B|C,D) \rangle$ abbreviates $\langle P(A,B|C,D) \rangle_{A,B}$ (the expectation values of $A$ and $B$ given particular $C$ and $D$).

To approximate the domain specific joint distributions $P(F, C, B|D)$ with a common $Q(F, C, B)$, we need to approximate all of the domain specific conditionals $P(C, B | F, D)$ with a common conditional $Q(C, B | F)$ and the domain specific marginals $P(F|D)$ by a common marginal $Q(F)$,
\begin{equation}
P(F,C,B|D) = P(C,B|F,D)P(F|D) \approx Q(C,B|F)Q(F).
\label{eq:covcon}
\end{equation}
Many state-of-the-art studies \cite{chen2018domain,muandet2013domain} attribute domain shift exclusively to the difference in marginals $P(F|D) \neq P(F|D')$ (i.e.\@~to covariate shift, Fig.\@~\ref{subfig:covariate}), while assuming conditionals to be stable across domains, $P(C,B|F,D)=P(C,B|F,D')$ (i.e.\@~no concept shift, Fig.\@~\ref{subfig:concept}).  
Alternatively, domain shift $P(F, C, B|D)$ can also be decomposed as:
\begin{equation}
P(F,C,B|D) = P(F|C,B,D)P(C,B|D),
\label{eq:labman}
\end{equation}
where changes between different domains in $P(C,B|D)$ are called \emph{label shift}, and in $P(F|C,B,D)$ are referred to as \emph{manifestation shift}. It is worth nothing that the two domain shift models in Eqs.\@~(\ref{eq:covcon}) and (\ref{eq:labman}) are mutually exclusive, as they rely on different assumptions about the causal direction between the class and the features in the real world.

The standard approach adopted by \cite{chen2018domain, matsuura2020domain}  of finding $\phi$ to address covariate shift by equalising the marginals across domains is through an explicit feature-level domain discriminator 
$s(F; \psi) \approx P(D|F)$,  by min-maxing the domain adversarial loss $\mathcal{L}_{dadv}$:
\begin{equation}
\begin{split}
\min_{\phi} \max_{\psi} \mathcal{L}_{dadv}(\phi, \psi), 
\end{split}
\label{eq:domain_loss}
\end{equation}
where $\mathcal{L}_{dadv}$ is defined as \cite{goodfellow2014generative}:
\begin{equation}
\begin{split}
\mathcal{L}_{dadv}(\phi, \psi) 
= \sum_I \mathbf{1}_{D(I)} . \log\left( s \left(f(I; \phi); \psi \right)\right),
\end{split}
\end{equation}
where $I$ ranges over the images in the training data, and $\mathbf{1}_D$ is the one-hot vector encoding the domain $D$ of image $I$. This minmax game enables $F=f(I;\phi)$ to learn features whose domain cannot be distinguished by any $s(F; \psi)$, leading to equality of domain marginals, 
\begin{equation}
\begin{split}
P(F | D  ) =&~  P(F|D') , \quad \forall D, D' \\ 
= &~ Q(F)
\end{split}
\end{equation}

However, as pointed out by recent studies \cite{li2018deep,zhao2020domain,scholkopf2012causal, janzing2010causal}, stability of conditionals across domains cannot be guaranteed. Any method aiming to achieve domain invariance needs to compensate for variation in conditionals $P(C, B | F, D)$. In other words, the domain discriminator $s(F;\psi)$ creates invariance on the sample space $\mathbf{I} \times \mathbf{D}$ but not on $\mathbf{I} \times \mathbf{C} \times \mathbf{B} \times \mathbf{D}$. Moreover, techniques proposed in recent studies \cite{li2018deep,zhao2020domain} are intended for classification and cannot be directly used to achieve generalised object detection.

\section{Proposed Method}
\label{sec:proposed_method}

We now describe how to approximate the domain-specific conditionals $P(C, B | F, D)=P(C, B | F, D')$ with a common $Q(C, B | F)$. In conjunction with Eq.~\eqref{eq:domain_loss} this will lead to a DG trainer.



To approximate the domain-specific conditional distribution $P(C,B|F,D)$ of each domain $D$ with a common distribution $Q(C, B | F)$, we minimise,

\begin{equation}
    \min_{\phi, \gamma, \beta} \sum\limits_{D} KL \left[  P(C, B | F, D) || Q(C, B | F ) \right]  
\label{eq:KL_div_init}
\end{equation}
\begin{equation}
= \min_{\phi, \gamma, \beta}  \sum\limits_{D} KL \left[ P(C |B, F, D) || Q(C | B, F ) \right] \\ 
+ \min_{\phi, \gamma, \beta} \sum\limits_{D} KL[P(B | F, D ) || Q(B | F ) ] 
\label{eq:KL_div_split}
\end{equation}
where $Q(C | B, F)$ and $Q(B | F)$ denote respectively the distributions associated with the instance level classifier and the bounding box predictor.  Through the minimisation of the two terms in Eq.~\eqref{eq:KL_div_split}, the overall system is encouraged to transform the input images into a feature space where the both domain-specific instance level classifier (first term) and the bounding box predictor (second term) will be domain invariant.
This implies that the optimisation in Eq.~\eqref{eq:KL_div_split} along with Eq.~\eqref{eq:domain_loss} will result in a domain generalised trained model.

\subsection{Classifier term}

The first term in Eq.~\eqref{eq:KL_div_split} can be further written as:

\begin{equation}
\min_{\phi, \gamma, \beta} \sum_D \langle \log \frac{P(C |B, F, D) }{Q(C | B, F)} \rangle 
\label{eq:KL_simplified_init}
\end{equation}

\begin{equation}
= \min_{\phi, \gamma, \beta} \left( \sum_D \langle \log P(C | B, F, D)  \rangle  - \sum_D \langle \log Q(C |B, F )   \rangle \right)
\label{eq:KL_simplified}
\end{equation}

\begin{equation}
= \min_{\phi, \gamma, \beta} \left( \sum_D -H(C | B, F, D)  +  \mathcal{L}_{cls}  \right)
\label{eq:KL_simplified_defs}
\end{equation}


The first term in Eq.~\eqref{eq:KL_simplified} is the sum of $|D|$ conditional negentropies $-H(C | B, F, D)$ over domains $D$. The second term in Eq.~\eqref{eq:KL_simplified} is known as the \textbf{classification loss}  $\mathcal{L}_{cls}$.   Note that many algorithms already exist for minimising $\mathcal{L}_{cls}$ so can be used as components of algorithms for minimising the whole of Eq.~\eqref{eq:KL_simplified}.

However such algorithms also need to minimise the conditional negentropy term, which is equivalent to maximising the conditional entropy $H(C |B, F, D)$.  As both minimisations involve the same parameters, iterative algorithms which act in parallel to minimise both terms may compete to pull the solution in opposite directions.

To see the effect of perfectly maximising the conditional entropy $H(C |B, F, D)$ alone in Eq.~\eqref{eq:KL_simplified} (i.e.\@~temporarily ignoring the competing objective $\mathcal{L}_{cls}$), we adapt the following theorem from \cite{zhao2020domain} for object detection: 

\textbf{Theorem 1:} Assuming all the object classes are equally likely, maximising $H(C |B, F, D)$ is equivalent to minimising the Jensen-Shannon divergence between the $|C|$ conditional distributions $P(B, F | C, D)$ across classes. The global optimum can be achieved if and only if:
\begin{equation}
\begin{split}
P(B, F  | C, D) = P(B, F | C', D) , \quad  \forall C,C' 
\label{eqn:thm-requirement}
\end{split}
\end{equation}
Even though this assumption can fail under a class imbalance scenario, balance can still be enforced through batch based biased sampling. 

\textbf{Proof:} We use the derivation from  supplementary material of \cite{zhao2020domain} and adapt it to the object detection setting by including the bounding box predictions $B$. The  information gain $G$ between random variables $L$ and $M$ is:
\begin{equation}
    G(L, M) = H(L) - H(L | M)
    \label{eq:info_gain}
\end{equation}
Let $L$ be a discrete random variable which can take the values in $\{1,\dots,|\mathbf{C}|\}$, then the relation between 
$KL$ and Jensen-Shannon ($JS$) divergence is given by \cite{endres2003new}:
\begin{equation}
    JS[P(M | L =1), P(M | L =2), \dots, P(M | L = |\mathbf{C}|) ] = \langle  KL[P(M | C) \Vert P(M)] \rangle_C
                   \label{eq:JS_KL_relation}
\end{equation}
By using the definition in Eq.~\eqref{eq:info_gain}, the class conditional negentropy can be expressed as 
\begin{equation}
    -H(C |B, F, D) = G(F, C, B|D) - H(C|D) 
    \label{eq:cls_conditional_expansion}
\end{equation}
The first term of Eq.~\eqref{eq:cls_conditional_expansion} can be expanded as 
\begin{align}
\nonumber
G(F, C, B|D) &= H(F, B|D) - H(F, B | C, D) \\
\nonumber
    & =\frac{1}{|\mathbf{C}|}\sum_C  \langle \log \frac{P(F, B | C, D )}{P(F, B, D)} \rangle \\
    & =\langle  KL(P(F, B | C, D) || P(F, B, D)) \rangle_C
    \label{eq:KL_divergence1}
\end{align}

From Eqs.~\eqref{eq:cls_conditional_expansion} and \eqref{eq:KL_divergence1}, it can be deduced that minimising the domain specific class conditional negentropy $-H(C |B, F,D)$ is equivalent to minimising the KL-divergence between $P(F, B | C, D)$ and $P(F, B, D)$. 
(Note that $H(C|D)$ is a ground truth constant which does not depend on our model parameters, so cannot be optimised.) 

By using Eq.~\eqref{eq:JS_KL_relation}, minimising the KL divergence in Eq.~\eqref{eq:KL_divergence1} is equivalent to minimising the Jensen-Shannon divergence:
\begin{equation}
\begin{split}
    JS[P(F, B | C = 1, D), P(F, B | C = 2, D), \dots, P(F, B | C = |\mathbf{C}|, D)]
\end{split}
\label{eq:JSD}
\end{equation}
whose global minimum occurs under the condition of Eq.~\eqref{eqn:thm-requirement} $\square$.



Eq.~\eqref{eqn:thm-requirement} shows that minimisation  
of class conditional negentropy alone results in domain specific instance level features with no discriminative ability to assign  correct class labels corresponding to the objects in the image $I$ -- making them completely useless for our actual goal of detection.   But in reality we do not wish to minimise this term alone, and instead use its hypothetical solution to pull against some existing method for minimising $\mathcal{L}_{cls}$ in Eq.~\eqref{eq:KL_simplified_defs}.

One way to solve Eq.~\eqref{eqn:thm-requirement} is to use adversarial learning, 
inspired by the minmax game approach proposed in \cite{zhao2020domain, goodfellow2014generative, gong2019twin}. We here introduce $|\mathbf{D}|$ new classifiers, $c(F; \gamma_D) $,
each parameterised by a different, domain-specific $\gamma_D$, and propose the following \textbf{domain specific adversarial} classification loss function $\mathcal{L}_{erc}$,
\begin{equation}
    \min_{\phi} \max_{\{ \gamma_D \}} \mathcal{L}_{erc}(\phi, {\gamma_D})
\end{equation}
where
\begin{equation}
\begin{split}
    \mathcal{L}_{erc}(\phi, \{\gamma_D\})  =  \sum_D \langle \log Q(C | B, F, D; \gamma_D)  \rangle
\end{split}
\end{equation}
and $Q(C | B, F, D; \gamma_D)$ is the $Q$ distribution with $\gamma_D$ replacing its usual $\gamma$.

\subsection{Bounding box term}

To optimise the second term in Eq.~\eqref{eq:KL_div_split}, we adopt a strategy previously used by \citet{chen2018domain} for DA. Minimising the KL divergence between the terms $P(B | F, D)$ and $Q(B | F)$ is equivalent to building a bounding box predictor independent of the domain label $D$. Using Bayes' theorem gives,
\begin{equation}
\begin{split}
    P(D | B, F)P(B | F) 
    = P(B | F, D)P(D | F)
    \end{split}
    \label{eq:domain_bayes}
\end{equation}
where $P(D | B, F)$ represents the instance level domain label predictor, $P(B | F, D)$ is the domain specific bounding box predictor, and $P(D | F)$ is the image level domain label predictor. From Eq.~\eqref{eq:domain_bayes}, we can observe that if there is a consistency between the image and instance level domain label predictor then the bounding box predictor will be invariant to domains, i.e.~$P(B | D, F) = P(B | F)$. 



The input $F[B]$ to the instance-level domain classifier will be the subset of image $I$ features, computed at locations within the bounding box $B$ by $f(I[B];\phi)$. The \textbf{instance level domain classification loss} $\mathcal{L}_{dins}$ employed at classifier $t(F[B];\tau)\approx P(D|F[B])$ is:
\begin{equation}
    \mathcal{L}_{dins}(\phi, \tau) = \sum\limits_{B \sqsubset  I} \langle \log(P(D | B, F))\rangle = \sum_D \sum\limits_{B \sqsubset  I} \mathbf{1}_D . \log \left( t \left( F[B] ; \tau\right)\right)
\label{eq:ins_DA_loss}
\end{equation}
where $B \sqsubset I$ ranges over the bounding boxes detected in image $I$.  As shown in Eq.~\eqref{eq:domain_bayes}, in order to achieve invariant bounding box prediction across domains, we need a \textbf{consistency regularisation loss} $\mathcal{L}_{cst}$  \cite{chen2018domain}:
\begin{equation}
    \mathcal{L}_{cst}(\phi, \beta) = \Vert  \langle  t( F[B] ; \tau) -  s(F; \psi) \rangle_{B \sqsubset I} \Vert_2 
    \label{eq.lcst}
\end{equation}
which is the $\ell^2$-norm of the average difference between instance and image domain discriminator vectors. Consequently, the main loss function that is to be used for training the system is given by:
\begin{equation}
\begin{split}
    \min_{\phi, \gamma} & \max_{\psi, \{ \gamma_D \},\tau} \mathcal{L}(\phi,  \gamma, \psi, \tau, \{\gamma_D\}, \beta)\\ 
    = & \mathcal{L}_{cls}(\phi, \gamma) + \mathcal{L}_{reg}(\phi, \beta) \\
    + & \alpha_1 \mathcal{L}_{dadv}(\phi, \psi) +  \alpha_2 \mathcal{L}_{dins}(\phi, \tau)\\
    + & \alpha_3 \mathcal{L}_{cst}(\phi, \beta) + \alpha_4 \mathcal{L}_{erc}(\phi, \{\gamma_D\})
\end{split}
\label{eq:main_loss}
\end{equation}
where $\alpha_1$, $\alpha_2$, $\alpha_3$, $\alpha_4$ represent the regularisation constants and $\mathcal{L}_{reg}$ the bounding box regression loss. The features learned via the maximisation of the domain specific classification loss $\mathcal{L}_{erc}$ can have a negative impact on the classifier $c(F; \gamma)$, and on the bounding box predictor $b(F; \beta)$, and thus can result in instability during the min-max optimisation of Eq.~\eqref{eq:main_loss}. To overcome this, following \cite{zhao2020domain}, we introduce $|\mathbf{D}|$ additional domain-specific classifiers, $c(F; \gamma'_D)$, with a \textbf{new cross-entropy loss} $\mathcal{L}_{cel}$:
\begin{equation}
    \mathcal{L}_{cel}(\phi, \{ \gamma_D^{'} \}) = -\sum_{D}  \langle \log Q (C | B, \bar{F}) | D ; \gamma'_D  \rangle - \sum_{D} \sum_{D' \neq D}   \langle \log Q  (C | B, F) | D' ; \bar{\gamma'}_D \rangle
    \label{eq:additional_classifier_loss}
\end{equation}
where the bar above $\bar{F}$ and $\bar{\gamma}_D$ indicates that the parameters are fixed during training. 

\subsection{Combined loss function}

The combined loss function $\mathcal{L}$ that we use to train the system is 
\begin{equation} 
\begin{split}
    \mathcal{L}(\phi, \beta,  \gamma, \psi, \tau,  \{\gamma_D\}, \{\gamma_D'\}) 
    = \mathcal{L}_{cls}(\phi, \gamma) &+ \mathcal{L}_{reg}(\phi, \beta) + \alpha_1 \mathcal{L}_{dadv}(\phi, \psi) \\
    &+ \alpha_2 \mathcal{L}_{dins}(\phi, \tau) + \alpha_3 \mathcal{L}_{cst}(\phi, \beta) \\ 
    &+ \alpha_4 \mathcal{L}_{erc}(\phi, \{\gamma_D\})   + \alpha_5 \mathcal{L}_{cel}(\phi, \{\gamma'_D\}) 
    \end{split}
\label{eq:final_main_loss}
\end{equation}
where $\alpha_5$ is the regularisation constant associated with the additional $|\mathbf{D}|$ domain specific classifiers $\gamma^{'}_D$. 

To minimize Eq.~\eqref{eq:final_main_loss} we require, 
\begin{equation} 
    \min_{\phi, \gamma, \beta, \{ \gamma_D'\}} \max_{\psi, \{ \gamma_D\}, \tau}  
    \mathcal{L}(\phi, \beta,  \gamma, \psi, \tau,  \{\gamma_D\}, \{\gamma_D'\}) 
\end{equation}

\subsection{Combined loss minimising DG trainer}


Since the additional $|\mathbf{D}|$ classifiers  $\bar{F}$ and $\bar{\gamma}_D$ are domain-specific, there is a high likelihood that they can prevent 
the learning of domain invariant features, which is against the goal of domain generalisation. But at the same time, the inclusion of the $\gamma'_D$ classifiers can help in overcoming the instability introduced by the $\gamma_D$ classifiers. Hence, an effective strategy for training these domain-specific classifiers is of crucial importance. To this end, we initially freeze $\phi$ and train each of the $\gamma'_D$ classifiers by using data from domain $D$. This step will aid the domain-specific classifiers to learn only domain invariant features. In the next step, we fix all $|\mathbf{D}|$ parameters $\gamma_D'$ and fine tune $\phi$ so that a sample $I_D$ from domain $D$ is classified accurately by all $\gamma^{'}_{D' \neq D}$. 

Algorithm \ref{alg:Alg1} shows a complete DG trainer to minimise $\mathcal{L}$ of Eq.~\eqref{eq:final_main_loss}, which optimises the main detector to achieve domain invariant bounding box prediction performance; which is further used for achieving class-conditional invariance.

\begin{algorithm}
\SetAlgoLined
\textbf{Input:} $|D|$ domain training datasets, $X_D$ 
\textbf{Input:} $\alpha_1$, $\alpha_2$, $\alpha_3$, $\alpha_4$, $\alpha_5$\\
\textbf{Input:} Feature and classifier functions $f, c, b$ \\
\textbf{Output:} trained model $\phi, \gamma, \beta$ \\
 \For{each epoch}{
  Sample data from each training dataset respectively\\
  Update $\phi, \beta, \gamma, \psi, \tau$ by optimising first five terms of Eq.~\eqref{eq:final_main_loss}\\
  \For{each domain $D$}{
  Sample data from the $D$-th dataset $X_D$\\
  Update $\gamma_D'$ by optimising the $\mathcal{L}_{cel}$ term of Eq.~\eqref{eq:final_main_loss}\\
  Update  $\phi, \gamma_D $ by optimising the $\mathcal{L}_{erc}$ term of Eq.~ \eqref{eq:final_main_loss}\\
  Sample data from datasets  $X_{D' \neq D}$ \\
  Update $\phi$ by optimising the $\mathcal{L}_{cel}$ term of Eq.~\eqref{eq:final_main_loss}\\
  }
 }
 \caption{Training strategy for domain generalised object detection.}
 
 \label{alg:Alg1}
\end{algorithm}

\begin{figure*}
    \begin{center}
    \includegraphics[width=1.0\textwidth]{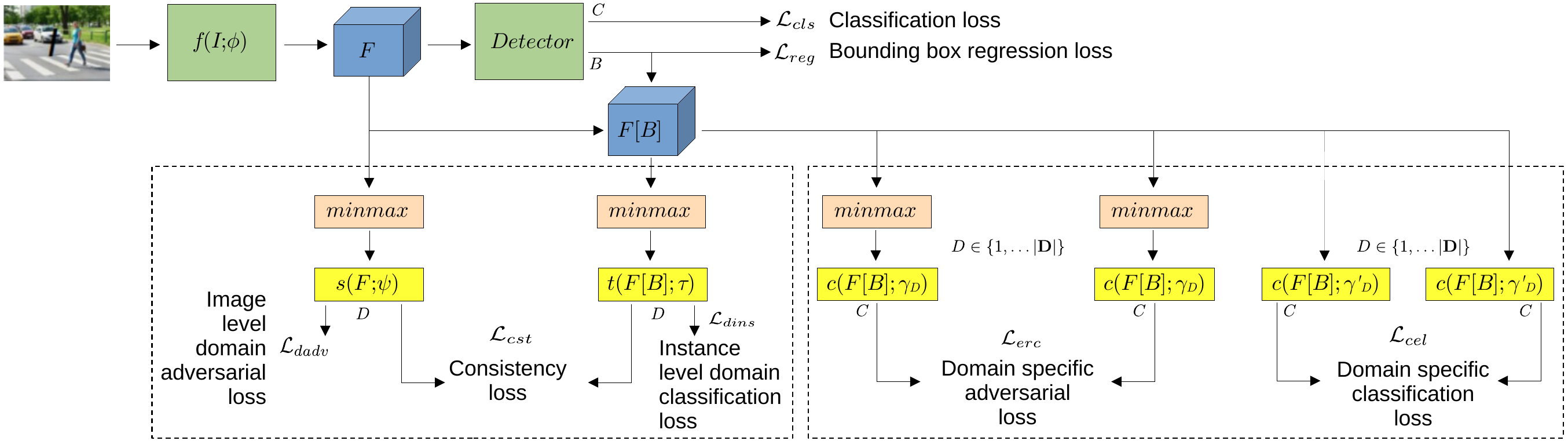}
    \end{center}
    \caption{Block diagram of the proposed approach.} 
    \label{fig:proposed_block_diagram}
\end{figure*}

A block diagram implementing the algorithm is given in Fig.~\ref{fig:proposed_block_diagram}. 
Compared against any generic object detector (one or two-stage), it introduces two new modules related to class-conditional invariance and bounding box invariance, which aid to optimise the feature extractor, in order to map the input images into feature vectors so that the detection is consistent across multiple domains.  The min-max optimisation is a form of adversarial learning which may be implemented by GRLs \cite{ganin2016domain}, which simply flip the sign of the error terms during loss backpropagation.


\section{Experimental Validation}
\label{sec:experimental_results}
\textbf{Datasets:} We demonstrate the generalisation ability of our approach on the following popular multi-source object detection datasets. As it is commonly practised in the state-of-the-art, we employ the official training set (source domain(s)) of each dataset for training and validation purposes, all the while the official validation set (target domain(s)) is reserved exclusively for testing, unless otherwise specified.

{\em Cityscapes} \cite{cordts2016cityscapes} contains detection-annotated urban street scenes. There are 2975 images (from 18 cities) for training and 500 validation images (from 3 cities). To conduct a fair comparison, we resize original images to 600 $\times$ 1200 resolution. \emph{Cityscapes} contains 8 classes: person, rider, car, truck, bus, train, motor, bike which are shared with other autonomous driving datasets. The number of instances across classes (for this and all other autonomous driving datasets used) is shown in Table \ref{tab:datasets}. {\em Foggy Cityscapes} is derived from Cityscapes via synthetic fog simulation \cite{foggy}.

{\em BDD10k}~\cite{bdd100k}: was proposed along with {\em BDD100k} for segmentation, and overlaps it significantly. Annotations are collected on six scene types, six weather conditions, and three times of  day. Unlike \emph{Foggy Cityscapes} \cite{foggy}, the fog in BDD10k is real. Train and validation sets have 7000 and 1000 images of 1280 $\times$ 720 pixel resolution respectively. Segmentation masks are converted into bounding boxes to fit the object detection setting. We select \emph{BDD10k} for our experiments as it is of commesurable size to other autonomous driving datasets used in this study, but use \emph{BDD100k} for a fair comparison with state-of-the-art (see \emph{Setups} below).

{\em ACDC}~\cite{acdc-dataset}:  Adverse Conditions Dataset with
Correspondences is an autonomous driving dataset targeted towards improving the performance of semantic segmentation task in adverse weather conditions. It has a total of 2006 images at 1920 $\times$ 1080 pixel resolution, which are uniformly spread across fog, nighttime, rain and snow conditions. 
We use the standard train validation split that comes with the dataset (1600 for training and 406 for validation).

{\em IDD}~\cite{idd-dataset}: Indian Driving Dataset (Parts I and II) aims for improved autonomous navigation and detection in unstructured environments. It has a total of 16063 images, split into 14027 and 2036 samples for training and validation respectively. Resolutions vary between 1280 $\times$ 720 and 1920 $\times$ 1080 pixels. They are captured from 182 driving sequences on Indian roads and are annotated with 34 classes.

{\em GWHD} \cite{david2021global,david2021dataset} contains annotated wheat head detections in images captured from multiple platforms and camera settings, with additional variation in wheat genotype, sowing density, growth stage and soil type. 6515 images (resolution: $1024 \times 1024$ pixels) are acquired across 47 sessions; each restricted to a single domain/farm. The training set has 18 domains with 3657 images; the validation set contains 11 different sessions with 1476 images; and the test set has 18 sessions with 1382 images. 
(There are small discrepancies in image counts reported in the paper and dataset; we use the splits provided by the dataset.) 


{\em URPC 2019 \cite{liu2020towards}} contains 3765 underwater images for training and 942 for validation with highest resolution $3840 \times 2160$ pixels, captured by a GoPro camera. Five classes are annotated: echinus (18490), starfish (5794), holothurian (5199),  scallop (6617) and  waterweeds (82), where numbers in parenthesis denote numbers of instances for each class. URPC2019 is formed via synthetic generation of 8 types of water quality as domains using style transfer.

\begin{table}[t]
    \caption{Number of instances of each class for all autonomous driving datasets} 
    \begin{center}
    \begin{tabular}{|c|c|c|c|c|c|c|c|c|}
    \hline
    \textbf{Dataset} & \textbf{person} & \textbf{rider} & \textbf{car} & \textbf{truck} & \textbf{bus} & \textbf{train} & \textbf{motorcycle} & \textbf{bicycle} \\
    \hline
    \emph{ACDC} & 2221 & 276 & 8522 & 643 & 217 & 258 & 210 & 482 \\ 
    \hline
    \emph{BDD10k} & 11140 & 516 & 82712 & 4266 & 1857 & 72 & 443 & 930 \\ 
    \hline
    \emph{Cityscapes} & 107065 & 11815 & 159110 & 2910 & 2415 & 970 & 4440 & 24520 \\ 
    \hline
    \emph{IDD} & 60617 & 73306 & 69929 & 20448 & 9540 & 15 & 83062 & 1905 \\ 
    \hline
    Total & 181043 & 85913 & 250344 & 28267 & 14031 & 1315 & 88155 & 27837\\
    \hline
    \end{tabular}
    \end{center}
    \label{tab:datasets}
\end{table}

\textbf{Setups:} The following experiments were performed to analyse generalisation ability of the proposed approach across multiple object detection architectures, multiple generalisation settings, and to assess their performance against existing state-of-the-art detectors equipped with generalisation capabilities.

\begin{enumerate}
\item {\em Multi-class, single target domain:} We evaluate the cross dataset generalisation ability of the object detectors. We use a leave-one-domain-out strategy across four autonomous driving datasets (\emph{ACDC}, \emph{BDD10K}, \emph{Cityscapes}, \emph{IDD}). The shift in this experiment is manifested in the form of new datasets which are captured by using completely different experiment protocols or geographical locations than the source domains. The results are presented in Table \ref{tab:cross_dataset_generalisation}, where {\em Single-best} indicates the best performing baseline model~(\fcos, \yolo, \frcnn) after training (and validating) independently with each of the source domains and testing with the validation set of the target domain. {\em Source-combined} indicates the combined use of all source domains in terms of training and validation sets and once again testing with the validation set of the target domain. {\em Ours} denotes the same setup as {\em Source-combined}, with the addition of the domain generalisation capacity to the tested detector. {\em Oracle-Train on Target} refers to training (and validating) using the target domain's training set, and to testing with the target domain validation set.
    
\item {\em Single-class, multiple target domains:} The \emph{GWHD} dataset enables evaluating the generalisation capacity of a model on a scenario with multiple target domains, where the shift in both the training and target domains stems from multiple factors such as acquisition location and equipment, soil type, wheat genotype, growth stage and sowing density (Table \ref{tab:domain_specific_analysis}). We include qualitative examples of successful (Fig.\@~\ref{fig:localise_gwhd}) and failed (Fig.\@~\ref{fig:GWHD_proposed_approach_failure}) detection cases, highlighting the strengths and weaknesses of the proposed approach. This is the only known public dataset aimed at evaluating domain generalisation performance within a single dataset.

\item {\em Ablation studies:} This experiment aims to measure the individual as well as combined effects on performance of the proposed bounding box and class conditional alignment modules. To this end, models relying on partial losses have been trained and evaluated on both single-class/multi-target domain and multi-class/single-target domain datasets across all three tested detectors (Table \ref{tab:GWHD_SCB2}). In addition to the aforementioned ablation experiments, we also provide a localisation performance comparison (Fig.~\ref{fig:improved_localisation}).

\item {\em Comparison with existing approaches:} Two experiments have been conducted to this end. The first compares the proposed \dgyolo against the reported IRM based DGYOLO \cite{liu2020towards} from the state-of-the-art, using the URPC2019 dataset (Table \ref{tab:URPC2019}). The second compares the proposed \dgfrcnn against the reported disentanglement based counterpart \cite{lin2021domain} from the state-of-the-art, using the \emph{BDD100k}, \emph{Cityscapes} and \emph{Foggy Cityscapes datasets} (Table \ref{tab:comparison_against_lin_etal}). Moreover, we highlight the limitation of the setup employed by \citet{lin2021domain}, where 3 datasets are used, but one of them is an order of magnitude larger than the others (\emph{BDD100k}) while one of them (\emph{Foggy Cityscapes}) is directly derived from the other (\emph{Cityscapes}). Instead, we propose using Setup-1 (presented in Table \ref{tab:cross_dataset_generalisation}) as a new, more extensive benchmark for the domain generalisation setting, as it employs 4 distinct datasets.
\end{enumerate}
    


\textbf{Network Architectures:}
We instantiate the general structure of Fig.\@~\ref{fig:proposed_block_diagram} using three different, neural network-based detectors, each extended with the new losses as additional neural network loss heads, and with GRLs to implement the adversarial learning (minmax). 
The three baseline models (\frcnn, \yolo, \fcos) share some structures but have various small differences, to optimise their individual performances. These optimisations are then retained into the DG versions. The number and sizes of layers used in our  instance classifiers differs slightly across  the models, to match the size of the feature maps from the baseline models. The training algorithms and training parameters are taken from the original model papers in order to accurately reproduce their baselines.

The image resolution used by \frcnn, \fcos and their DG counterparts for autonomous driving dataset is 1200 $\times$ 600, in line with DA works on these datasets, and the original image resolution of 1024 $\times$ 1024 is used for \emph{GWHD}. The image resolution used in all \yolo experiments is 640 $\times$ 640, the only native resolution supported by this detector.
Based on previous empirical optimisation with FRCNN, \cite{seemakurthy2023aaai}, the regularisation constants for all models were initially set to $\alpha_1 = 0.5$, $\alpha_2 = 0.5$, $\alpha_3 = 0.5$, $\alpha_4 = 0.05$, and $\alpha_5 = 0.0001$, with parameters $\alpha_3$ and $\alpha_4$ (corresponding to the consistency regulariser and domain specific adversarial loss) hand-optimised empirically for each detector. We note any differences to these default parameters when discussing individual detectors. 



\begin{figure*}
    \begin{center}
    \includegraphics[width=1.0\textwidth]{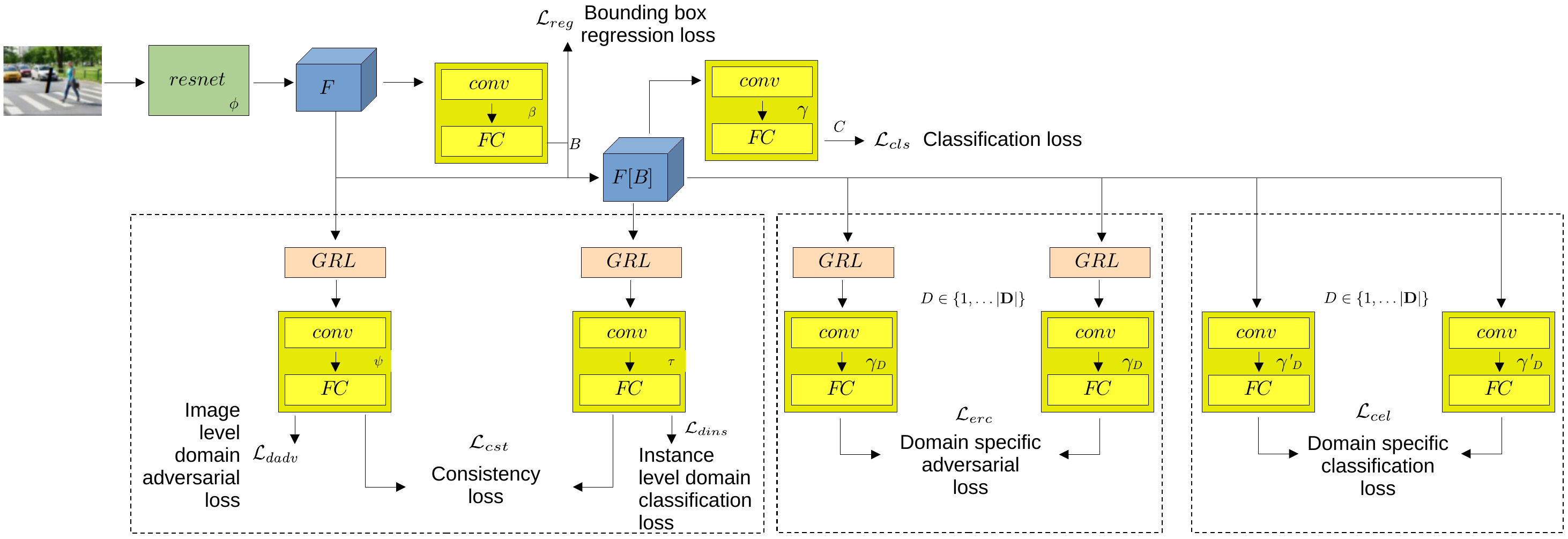}
    \end{center}
    \caption{Block diagram of the proposed approach, instantiated as FRCNN with neural network heads.}
    \label{fig:proposed_block_diagram_frcnn}
\end{figure*}

{\em FRCNN} \cite{ren2015faster} is a two-stage detector, instantiated in the DG framework as in fig.~\ref{fig:proposed_block_diagram_frcnn}.  We here use a ResNet50-FPN backbone network  initialised with pre-trained pretrained COCO weights \cite{lin2014microsoft} for the autonomous driving experiments, and ImageNet weights \cite{deng2009imagenet} for \emph{GWHD}. The feature extractor $f$ is the backbone, region proposal network and ROI pooling layer of the \frcnn network.  $b$ is the bounding box regressor and $c$ is the instance-level classifier components of the \frcnn.  The output $F$ of the backbone network is fed as input to domain adversarial network $s(F;\psi)$ while the output of ROI Pooling layer of the \frcnn detector is fed as an input to instance level domain classifier $t(F[B];\tau)$, $2|D|$ domain specific classifiers ($\gamma_D$ and $\gamma_D'$).  All terms in the loss function described in Eq.~\eqref{eq:final_main_loss} correspond to either a domain or object classifier. We use cross-entropy to train each of these classifier modules.  The resulting network is shown in Fig.~\ref{fig:proposed_block_diagram_frcnn}.   Training: We used early stopping with a patience of 10 epochs. AdamW (weight decay = $0.0005$, learning rate = $0.001$, batchsize=$2$) has been used as optimiser while training with \emph{GWHD} and Stochastic Gradient Descent (SGD) (weight decay = 0.0005, momentum=0.9, learning rate=\num{2e-3}, batchsize=$2$) has been used for the autnomous driving experiments. Different to the default value, in \emph{GWHD} experiments $\alpha_4=0.075$. \frcnn has a total of 41.3 M parameters while the \dgfrcnn uses 48.1 M and 67.8 M number of parameters for training on the autonomous driving and \emph{GWHD} datasets, respectively (the number of inference parameters remains the same as for the baseline architecture). 


\begin{figure*}
    \begin{center}
    \includegraphics[width=1.0\textwidth]{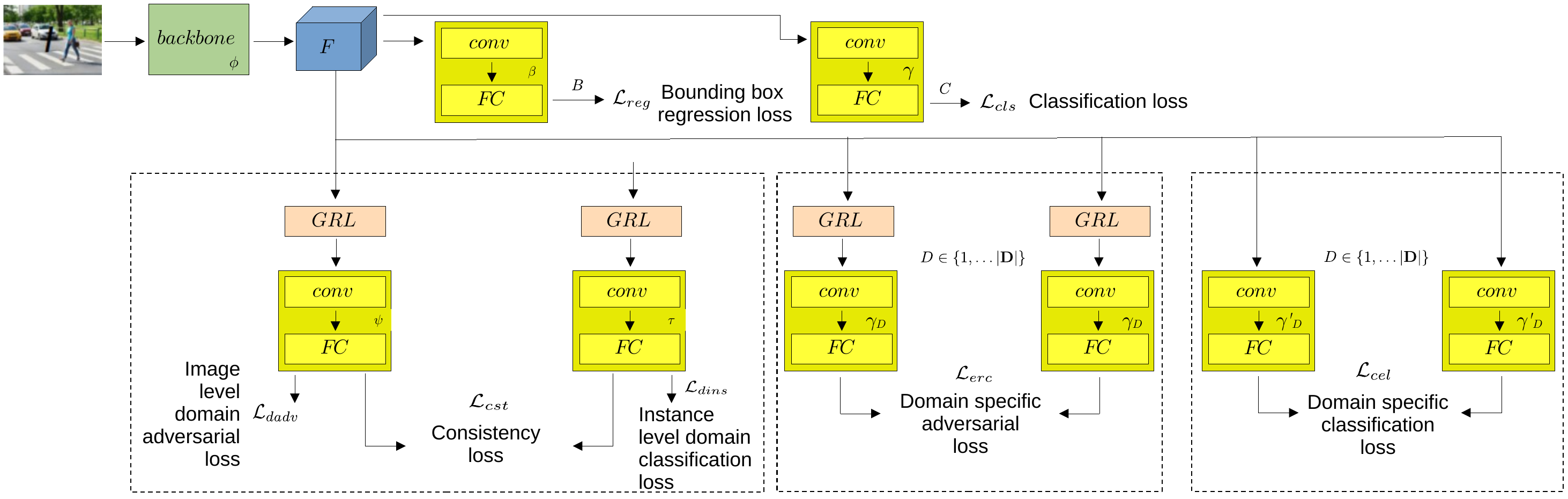}
    \end{center}
    \caption{The proposed approach instantiated as single stage detector with neural networks.  
    }
    \label{fig:proposed_block_diagram_singlestage}
\end{figure*}

{\em \yolo} \cite{redmon2018yolov3,redmon2016you} is a single-stage detector, which means that $B$ and $C$ are assumed to be independent given $F$ and $D$, i.e. $P(B,C|F,D)=P(B|F,D)P(C|F,D)$, resulting in the network structure of Fig.~\ref{fig:proposed_block_diagram_singlestage}   We here use a DarkNet-53 backbone network initialise with COCO pretrained weights. $F$ is the top level of the backbone, evaluated at a set of anchor points.
The output of the $F$ for \yolo are the features at multiple different scales. Anchor boxes of three different aspect ratios are regressed at every spatial/grid location of these multi-scale features. Training used SGD  (learning rate = 0.1, weight decay = 0.0005, momentum = 0.9, batchsize=8) for autonomous driving datasets, and AdamW (learning rate = 0.001, weight decay = 0.0005) for \emph{GWHD} and \emph{URPC}. Different to default, $\alpha_3=0.8$ in autonomous driving experiments, and $\alpha_3=0.5$ and $\alpha_4=0.055$ for \emph{GWHD}.
\yolo has 61.5 M parameters while \dgyolo trains with 63.4 M and 110 M parameters for driving and \emph{GWHD} datasets, respectively.

{\em \fcos} \cite{tian2019fcos} is another single-stage detector, so shares the same network structure as   Fig.~\ref{fig:proposed_block_diagram_singlestage}. The backbone used is ResNet50.  $F$ is the  the top {\em three} layers of the backbone network, plus five further layers linked from them and to each other, with the bounding box and classification heads swept over each of these.  (An additional centerness head is also used to weight which detections to report).  
We used early stopping with a patience of 10 epochs. To replicate original dataset baselines, AdamW (weight decay = 0.0001, learning rate = 0.0001, batchsize=8) is used as the optimiser while training with \emph{GWHD} and Stochastic Gradient Descent (SGD) (weight decay = 0.0001, momentum=0.9, learning rate=\num{0.01}, batchsize=8) has been used for autonomous driving datasets.  The feature extractor $F$ is the top level of the backbone, evaluated at a set of anchor points.  \fcos has a total of 32.1 M parameters while the \dgfcos trains with 56.9 M and 109 M parameters for driving and \emph{GWHD} datasets, respectively.



\textbf{Implementation:} Experiments ran on a 24GB, 10,496 core NVidia RTX 3090 GPU. Each epoch, including validation, lasted around 20 minutes.   
Experiments were coded in PyTorch 2.1.2 using CUDA 12.1, by extending existing \frcnn and \fcos models from the Torchvision library, and \yolo from \url{github.com/eriklindernoren/PyTorch-YOLOv3} (same as \cite{liu2020towards}), with DG heads. Code is GPLv3 open source and made available, together with models and documentation to replicate all experiments.\footnote{github repo will be made public upon acceptance}

\textbf{Metrics}: Standard practice is used to  measure performance of object detection, and report \emph{mean Average Precision ($\operatorname{mAP}$)} at $50\%$ \emph{Intersection over Union ($\operatorname{IoU}$)} -- $\operatorname{mAP@50}$ \cite{everingham2010pascal} ($\operatorname{mAP}$ for short) as our main performance metric.

Every output from the detector is assigned a confidence score $p\in [0,1]$, and, for every class $C \in \mathbf{C}$, \emph{precision} and \emph{recall} values are calculated at every confidence level $T$, by dividing the detector outputs into \emph{true positives} ($\operatorname{TP}_{C,T}$) and \emph{false positives} ($\operatorname{FP}_{C,T}$) according to IoU with the ground truth. Detector outputs with $p\geq T$ and overlap with the ground truth of the same class higher or equal to $\operatorname{IoU}=50\%$ are deemed ($\operatorname{TP}_{C,T}$), and all others ($\operatorname{FP}_{C,T}$). Any ground truth annotation which was not matched with a detector output is deemed a \emph{false negative} ($\operatorname{FN}_{C,T}$). Finally, for each class $C$ and at each confidence level $T$, \emph{precision} is defined as a fraction of correct detector outputs:
\begin{equation}
p_{C,T}=\frac{\operatorname{TP}_{C,T}}{\operatorname{TP}_{C,T}+\operatorname{FP}_{C,T}},
\end{equation}
and \emph{recall} as a fraction of ground truth objects which were correctly detected:
\begin{equation}
r_{C,T}=\frac{\operatorname{TP}_{C,T}}{\operatorname{TP}_{C,T}+\operatorname{FN}_{C,T}}.
\end{equation}
This allows us to further define \emph{precision at $r$}, as the precision value corresponding to a certain recall level $r$:
\begin{equation}
    p_{C}(r) = p_{C,T} \textnormal{ s.t. } r_{C,T} = r,
\end{equation}
as well as \emph{interpolated precision}, corresponding to the maximum value of precision at every recall level, and required to remove the sensitivity of precision and recall measures to ranking differences (i.e.\@~the sample order during evaluation):
\begin{equation}
    \tilde{p}_C(r) = \max_{r'\geq r} p_C(r').
\end{equation}
The pairs of (interpolated) precision and recall values at different confidence scores are used to calculate precision-recall curves for each class. \emph{Average Precision} for class $C$ ($\operatorname{AP}_C$) summarises the shape of the precision-recall curve for each class across all confidence levels (in our case, calculated as an 101-point approximation), corresponding to the area under the curve (calculated at $\operatorname{IoU}=50\%$):
\begin{equation}
    \operatorname{AP}_C = \frac{1}{101}\sum_{i=0 \atop r=0.01i}^{100} \tilde{p}_C(r).
\end{equation}
Finally, $\operatorname{mAP}$ is calculated as the mean of $\operatorname{AP}_C$ for all classes $C$ present in the dataset:
\begin{equation}
\operatorname{mAP} = \frac{1}{|\mathbf{C}|}\sum_{C\in\mathbf{C}}\operatorname{AP}_C.
\end{equation}

\section{Results} 
\label{sec:results}

\textbf{Multi-class, single target domain.} Results using multiple autonomous driving datasets, are presented in Table\@~\ref{tab:cross_dataset_generalisation}. It can be seen that addition of our proposed alignment modules during training gives  consistent performance improvements over both the \emph{Single-best} and \emph{Source-combined} (except for the \fcos detector in the case of (C, I, A) $\rightarrow$ B; however in this case the difference between \emph{Source-combined}, \emph{Oracle} and our proposed approach is less than a single percentage point). We also surpass the \emph{Oracle} in 5 distinct experiments, most notably in 3 out of 4 experiments with \yolo. While per-class improvement is not achieved in all cases, our proposal improves performance on half or more classes present in datasets in 10 our of 12 cases. (Results for the \emph{train} class are poor across all settings, as it is severely underrepresented across all datasets.)

\begin{table*}[!ht]
\caption{Cross dataset generalisation performance on \emph{ACDC} (A), \emph{BDD10K} (B), \emph{Cityscapes} (C), \emph{IDD} (I). The best AP and mAP in each setting (excluding \emph{Oracle}) is highlighted in bold. {\em (X, Y, Z) $\rightarrow$ W}: X, Y, Z refer to the source domains, and W denotes the target domain.
}
\begin{center}
\small{
\begin{tabular}{c|c|cccccccc|c}
    \hline
    DG Setting & Methods & person & rider & car & truck & bus & train & motor & bike & mAP \\
    \hline
    \multicolumn{11}{c}{\textbf{\fcos}}\\
    \hline
    \multirow{5}{*}{(A, B, C) $\rightarrow$ I} & Single-best & 24.48 & 22.02 & 48.98 & 28.13 & 16.53 & 0.00 &  21.01 & 1.99 & 20.39 \\
                                & Source-combined & \textbf{29.32} & \textbf{29.01} & \textbf{50.75} & 29.40 & 14.38 & 0.00 & \textbf{31.35} & 13.01 & 24.65   \\
                                & \dgfcos (Ours) & 28.73 & 28.65 & 47.96 & \textbf{29.63} & \textbf{18.59} & 0.00 & 30.93 & \textbf{14.74} & \textbf{24.90}   \\
                                \cline{2-11}
                                & Oracle-Train on Target & 44.45 & 50.35 & 62.81 & 57.95 & 53.88 & 0.00 & 52.79 & 28.77 & 43.87  \\
                                \cline{2-11}
    \hline
    \hline
    
    \multirow{5}{*}{(B, C, I) $\rightarrow$ A} & Single-best & 35.56 & 16.82& 65.29& 21.15& 05.43& 0.00& 23.94& 08.19&22.05   \\
                             & Source-combined & 41.71 & \textbf{21.14} & \textbf{73.96} & 34.30 & \textbf{13.30} & 17.53 & 28.95 & 12.23 & 30.39\\
                             & \dgfcos (Ours) & \textbf{44.08} & 18.24 & 73.37 & \textbf{36.17} & 13.13 & \textbf{19.73} & \textbf{29.35} & \textbf{13.39} & \textbf{30.93}   \\
                                \cline{2-11}
                             & Oracle-Train on Target & 36.08 & 10.55 & 65.48 & 19.17 & 10.63 & 24.03 & 12.91 & 04.49 & 22.92  \\ 
                                \cline{2-11}
    \hline
    \hline
    \multirow{5}{*}{(C, I, A) $\rightarrow$ B} & Single-best  & 51.60 & 29.06 & 59.79 & 14.84 & 19.25 & 0.00 & 24.89 & 25.41 & 28.10 \\
                             & Source-combined & 54.48 & 37.98 & 62.70 & \textbf{18.64} & 21.99 & \textbf{7.43} & \textbf{33.48} & 25.27 & \textbf{32.75} \\
                             & \dgfcos (Ours) & \textbf{54.81} & \textbf{40.00} & \textbf{62.91} & 16.83 & \textbf{24.17} & 0.33 & 30.50 & \textbf{27.23} & 32.10   \\
                                \cline{2-11}
                             & Oracle-Train on Target & 51.36 & 29.91 & 67.19 & 30.83 & 31.32 & 0.65 & 22.66 & 23.69 & 32.20   \\
                                \cline{2-11}
    \hline
        \multirow{5}{*}{(I, A, B) $\rightarrow$ C} & Single-best & 43.41 & 41.23 & 63.19 & 25.26 & 43.99 & 0.00 & \textbf{35.46} & 30.20 & 35.34  \\
                             & Source-combined & \textbf{47.57} & \textbf{48.60} & 69.02 & 36.19 & \textbf{48.59} & \textbf{13.68} & 34.11 & 34.54 & 41.54   \\
                             & \dgfcos (Ours) & 46.91 & 45.51 & \textbf{72.50} & \textbf{37.32} & 52.21 & 11.93 & 32.55 & \textbf{37.26} & \textbf{42.02}   \\
                                \cline{2-11}
                             & Oracle-Train on Target & 50.08 & 51.66 & 68.46 & 23.63 & 51.72 & 14.18 & 35.60 & 43.70 & 42.38  \\
                                \cline{2-11}

    \hline

    \multicolumn{11}{c}{\textbf{\yolo}} \\
    \hline
    \multirow{5}{*}{(A, B, C) $\rightarrow$ I} & Single-best & 9.10& 1.80& \textbf{40.73}& 10.12&  6.74& 0.00& 2.77& 2.13 & 9.17 \\
                                & Source-combined & 15.87&  \textbf{9.34}& 40.11& 13.35&  9.92&  0.00&  13.49&  9.28& 13.92 \\
                                & \dgyolo (Ours) & \textbf{15.97} & 9.09 & 40.34 & \textbf{14.32} & \textbf{10.92} & 0.00 & \textbf{13.63} & \textbf{11.14} & \textbf{14.43}   \\
                                \cline{2-11}
                                & Oracle-Train on Target & 26.39& 37.92& 53.05&  42.07& 43.51& 0.00& 48.91& 16.49& 33.54 \\
                                \cline{2-11}
    \hline
    \multirow{5}{*}{(B, C, I) $\rightarrow$ A} & Single-best & 13.55&  10.50&  50.14&  10.67&  07.53&  0.00&  \textbf{30.88}&  4.75&  16.00 \\
                                & Source-combined & \textbf{23.04} & 17.22 & 62.32 & 21.03 & 16.49 & 0.00 & 18.54 & \textbf{10.99}  & 21.20   \\
                                & \dgyolo (Ours) & 20.47 & \textbf{20.64} & \textbf{65.81} & \textbf{30.45} & \textbf{18.75} & 0.00 & 11.31  & 10.91 &  \textbf{22.29} \\
                                \cline{2-11}
                                & Oracle-Train on Target & 12.26& 2.97& 54.12& 8.02& 5.94& 7.78& 9.62& 6.85& 13.44   \\
                                \cline{2-11}
    \hline

    \multirow{5}{*}{(C, I, A) $\rightarrow$ B} & Single-best & 29.90&  11.54&  47.31&  7.39&  \textbf{12.05}&  0.00&  28.71&  7.08 & 18.00 \\
                                & Source-combined & \textbf{40.19} & 25.41& \textbf{54.10}& 08.04& 11.44& 0.00& 28.52& 14.10& 22.72  \\
                                & \dgyolo (Ours) & 40.00 & \textbf{31.37} & 53.68 & \textbf{8.85} & 9.88 & 0.00 & \textbf{33.36} & \textbf{15.45} & \textbf{24.07}   \\
                                \cline{2-11}
                                & Oracle-Train on Target & 23.21& 08.91& 53.07& 16.65& 17.23& 0.00& 10.89& 08.25& 17.28   \\
                                \cline{2-11}
    \hline
    \multirow{5}{*}{(I, A, B) $\rightarrow$ C} & Single-best & 18.24& 11.48& 44.58& 15.21& 34.49& 0.00& 26.20& 07.44 &19.70 \\
                                & Source-combined & \textbf{24.90}& 24.16& 54.14& 28.81& \textbf{43.97}& 0.00& 30.24& \textbf{17.98} & 28.02  \\
                                & \dgyolo (Ours) & 22.82 & \textbf{24.69} & \textbf{56.29} & \textbf{32.32} & 43.16 & 0.00 &  \textbf{32.15} & 13.95 & \textbf{28.17}   \\
                                \cline{2-11}
                                & Oracle-Train on Target & 27.82& 29.93& 56.23& 15.43& 32.31& 0.00& 19.05& 26.17& 25.87  \\
                                \cline{2-11}
    \hline
        \multicolumn{11}{c}{\textbf{\frcnn}} \\
    \hline
    \multirow{5}{*}{(A, B, C) $\rightarrow$ I} & Single-best & 31.61 & 26.95 & 57.34 & 32.96 & 29.67 & 0.00 & 30.42 & 16.77 & 28.21  \\
                                &Source-combined & \textbf{35.17} & \textbf{30.68} & 56.84 & 34.21 & 28.80 & 0.00 & \textbf{35.74} & 18.79 & 30.03  \\
                                & \dgfrcnn (Ours) & 33.64 & 29.65 & \textbf{57.66} & \textbf{36.53} & \textbf{29.84} & 0.00 & 34.42 & \textbf{21.27} & \textbf{30.38}  \\
                                \cline{2-11}
                                & Oracle-Train on Target & 44.20 & 50.81 & 64.44 & 56.04 & 52.32 & 0.00 & 58.23 & 28.92 & 44.37 \\
                                \cline{2-11}
    \hline
    
    \multirow{5}{*}{(B, C, I) $\rightarrow$ A} & Single-best & 33.59 & 18.86 & 71.61 & 26.06 & 14.08 & \textbf{35.63} & 27.93 & 10.71 & 29.81   \\
                                & Source-combined & 41.09 & 18.51 & 72.36  & \textbf{43.39} & 24.77 & 26.50 & 38.81  & \textbf{12.54} & 34.75  \\
                                & \dgfrcnn (Ours) & \textbf{42.03} & \textbf{20.61} & \textbf{74.57} & 43.18 & \textbf{25.90} & 29.58 & \textbf{40.68} & 11.42 & \textbf{36.00}  \\
                                \cline{2-11}
                                & Oracle-Train on Target & 40.83 & 28.75 & 73.32 & 37.77 & 32.51 & 49.39 & 38.56 & 15.57 & 39.59  \\ 
                                \cline{2-11}
    \hline
    
    \multirow{5}{*}{(C, I, A) $\rightarrow$ B} & Single-best & 53.39 & 26.40 & 61.71 & 19.43 & 26.62 & 0.00 & 35.50 & 17.90 & 30.14 \\
                                & Source-combined & 58.59 & 45.75 & 65.39 & \textbf{25.74} &  \textbf{27.74} & 8.70 & 37.38 & 16.32 & 34.72  \\
                                & \dgfrcnn (Ours) & \textbf{58.67} & \textbf{47.00} & \textbf{65.53} & 24.44 & 25.90 & \textbf{9.91} & \textbf{40.83} & \textbf{19.10} & \textbf{35.31}  \\
                                \cline{2-11}
                                & Oracle-Train on Target & 55.19 & 33.55 & 69.43 & 38.04 & 37.74 & 0.00 & 32.54 & 21.71 & 36.02   \\
                                \cline{2-11}
    \hline
    \multirow{5}{*}{(I, A, B) $\rightarrow$ C} & Single-best & 39.76 & 30.74 & 61.55 & 35.32 & 40.29 & \textbf{37.70} & 26.95 & 32.17 & 38.06  \\
                                & Source-combined & 48.80 & \textbf{43.07} & 69.68 & \textbf{42.73} & 55.33 & 2.48 & \textbf{44.23} & 34.22 & 42.57   \\
                                & \dgfrcnn (Ours) & \textbf{49.46} & 42.76 & \textbf{70.26} & 42.54 & \textbf{56.01} & 12.62 & 44.16 & \textbf{37.11} & \textbf{44.36}   \\
                                \cline{2-11}
                                & Oracle-Train on Target & 54.51 & 51.10 & 73.84 & 42.01 & 58.75 & 47.72 & 44.06 & 47.33 & 52.42  \\ 
                                \cline{2-11}
    \hline
\end{tabular}}
\end{center}
\label{tab:cross_dataset_generalisation}
\end{table*}

\begin{figure}[t]
    \centering
    \begin{tabular}{ccccc}
        \includegraphics[width=80px, height=80px]{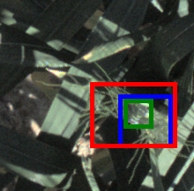} &  
        \includegraphics[width=80px, height=80px]{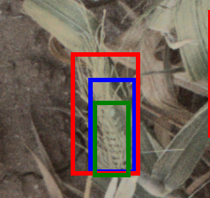} & 
        \includegraphics[width=80px, height=80px]{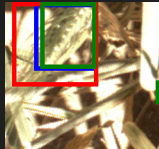} & 
        \includegraphics[width=80px, height=80px]{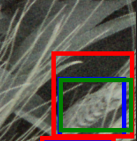} & 
        \includegraphics[width=80px, height=80px]{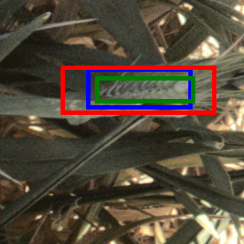} \\
        \includegraphics[width=80px, height=80px]{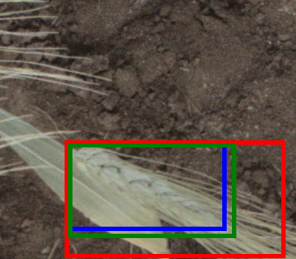} &  
        \includegraphics[width=80px, height=80px]{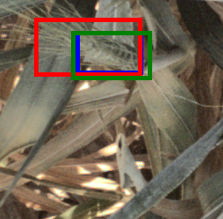} & 
        \includegraphics[width=80px, height=80px]{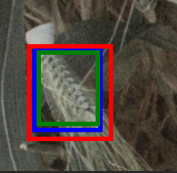} & 
        \includegraphics[width=80px, height=80px]{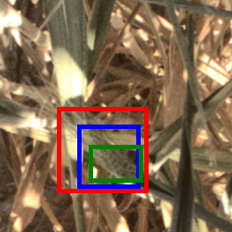} & 
        \includegraphics[width=80px, height=80px]{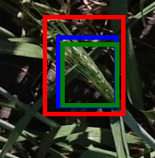} \\
        \includegraphics[width=80px, height=80px]{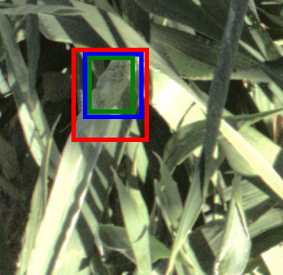} &  
        \includegraphics[width=80px, height=80px]{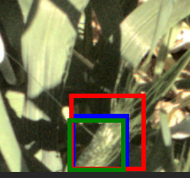} & 
        \includegraphics[width=80px, height=80px]{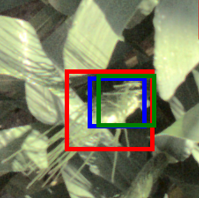} & 
        \includegraphics[width=80px, height=80px]{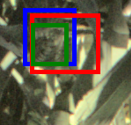} & 
        \includegraphics[width=80px, height=80px]{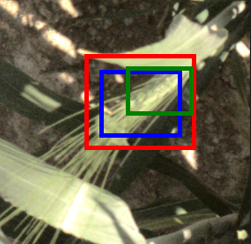} \\
    \end{tabular}
    \caption{Localisation performance on \emph{GWHD}. Output of: First row: \fcos vs \dgfcos. Second row: \yolo vs \dgyolo. Third row: \frcnn vs \dgfrcnn. Ground truth \crule[c]{0.5cm}{0.25cm}, Baseline \crule[d1]{0.5cm}{0.25cm} and DG framework \crule[d3]{0.5cm}{0.25cm}.}
    \label{fig:localise_gwhd}
\end{figure}

\begin{figure}[t]
    \centering
    \begin{tabular}{cccc}
    \includegraphics[width=80px,height=80px]{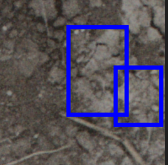} &  
    \includegraphics[width=80px,height=80px]{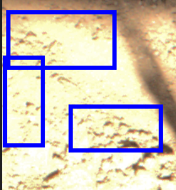} &
    \includegraphics[width=80px,height=80px]{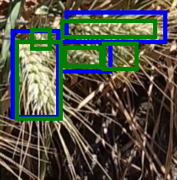} &  
    \includegraphics[width=80px,height=80px]{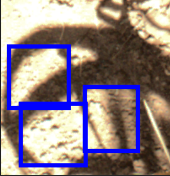} \\
    \includegraphics[width=80px,height=80px]{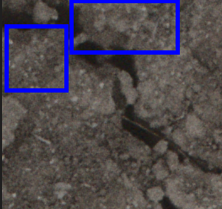} &  
    \includegraphics[width=80px,height=80px]{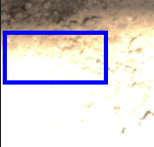} &
    \includegraphics[width=80px,height=80px]{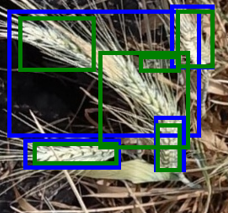} &  
    \includegraphics[width=80px,height=80px]{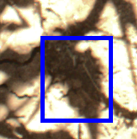} \\
    \includegraphics[width=80px,height=80px]{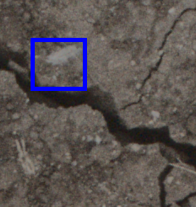} &  
    \includegraphics[width=80px,height=80px]{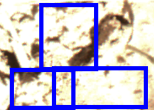} &
    \includegraphics[width=80px,height=80px]{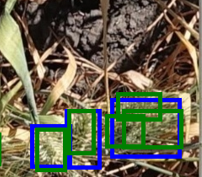} &  
    \includegraphics[width=80px,height=80px]{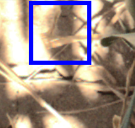} \\
    \end{tabular}
    \caption{Failure case analysis for \emph{GWHD}  on: First row: \dgfcos. Second row: \dgyolo. Third row: \dgfrcnn. Colour codes:~Predictions \crule[d3]{0.5cm}{0.25cm}, Ground Truth \crule[c]{0.5cm}{0.25cm}.}
    \label{fig:GWHD_proposed_approach_failure}
\end{figure}

\textbf{Single-class, multiple target domains.} This experiment allows us to examine the effect of our proposed method in the case where a trained model is deployed on multiple target domains at once. The results, detailed in Table\@~\ref{tab:domain_specific_analysis}, show that the addition of our modules on top of all three baseline architectures (\frcnn, \fcos, \yolo trained on all available training data, equivalent to the \emph{Source-combined} setting in the multi-class experiment) increases overall performance. It can also be seen that improvement across the domains for  \fcos and \yolo (16/18 domains) is more consistent than for \frcnn (13/18 domains).
This may indicate that our method is more beneficial to the performance of single-stage detector architectures than that of two-stage architectures, possibly because the instance-level alignment still acts on the features of the whole image (unlike with two-stage approaches, where instance-level alignment acts only on the features contained within the object bounding box). We also show example detector outputs both for successful detections (Fig.\@~\ref{fig:localise_gwhd}) and failure cases (Fig.\@~\ref{fig:GWHD_proposed_approach_failure}). Fig.\@~\ref{fig:localise_gwhd} indicates at the improved localisation performance across all baseline approaches (discussed in the next paragraph). The examples in Fig.\@~\ref{fig:GWHD_proposed_approach_failure} show that all approaches produce false detections when the background (twigs, textures and shadows) resemble the target class (wheat head), and struggle in dense regions with many objects and mutual occlusions.

\begin{table*}[tb]
    \caption{$\operatorname{mAP}$ performance improvement per domain, of the proposed framework over baselines from different networks on GWHD.}
    \begin{center}
    \small{
    \begin{tabular}{|c|c|c|c|c|c|c|c|}
    \hline
     Domain ID & \# of Images & \frcnn & \dgfrcnn &  \fcos & \dgfcos & \yolo & \dgyolo \\
     \hline
     UQ_7   & 17 & 66.45 & 75.85 ({\color{c}+9.40}) & 71.90 & 77.47 ({\color{c}+5.57})& 56.10 & 59.25 ({\color{c}+3.15})  \\
     \hline
     UQ_8   & 41 & 48.04 & 52.88 ({\color{c}+4.84}) & 49.36 & 52.25 ({\color{c}+2.89}) & 35.44 & 38.10 ({\color{c}+2.66})\\
     \hline
     UQ_9   & 33 & 51.08 & 55.70 ({\color{c}+4.62}) & 50.75 & 56.09 ({\color{c}+5.34}) & 37.09 & 39.29 ({\color{c}+2.2})\\
     \hline
     UQ_10   & 106 & 52.95 & 57.35 ({\color{c}+4.40}) & 53.81 & 54.98 ({\color{c}+1.17}) & 35.06 & 38.92 ({\color{c}+3.86})\\
     \hline
     UQ_11   & 84 & 38.71 & 43.40 ({\color{c}+4.69})& 42.83 & 43.96 ({\color{c}+1.13})& 25.87 & 28.80 ({\color{c}+2.93}) \\
     \hline
     Terraref_1  & 144 & 20.86 & 18.02 ({\color{red}-2.84})& 17.04 & 12.17 ({\color{red}-4.87})& 8.97 & 4.50 ({\color{red}-4.47}) \\
     \hline
     Terraref_2   & 106 & 3.52 & 4.50 ({\color{c}+0.98})& 5.81 & 9.21 ({\color{c}+3.4}) & 9.48 & 12.50 ({\color{c}+3.02}) \\
     \hline
     KSU_1   & 100 & 62.75 & 62.45 ({\color{red}-0.30})& 61.17 & 63.12 ({\color{c}+1.95}) & 52.20 & 58.08 ({\color{c}+5.88})\\
     \hline
     KSU_2   & 100 & 67.02 & 68.08 ({\color{c}+1.06})& 65.54 & 70.12 ({\color{c}+4.58}) & 50.05 & 52.20 ({\color{c}+2.15})\\
     \hline
     KSU_3   & 95 & 64.60 & 64.40 ({\color{red}-0.20}) & 58.81 & 63.91 ({\color{c}+5.1}) & 51.59 & 56.78 ({\color{c}+5.19}) \\
     \hline
     KSU_4   &  60 & 58.15 & 56.93 ({\color{red}-1.22}) & 55.37 & 61.08 ({\color{c}+5.71}) & 44.56 & 42.10 ({\color{red}-2.46})\\
     \hline
     CIMMYT_1  & 69 & 54.83 & 54.65 ({\color{red}-0.18}) & 50.89 & 54.23 ({\color{c}+3.34})& 31.00 & 37.12 ({\color{c}+6.12})\\
     \hline
     CIMMYT_2   & 77 & 68.13 & 71.70 ({\color{c}+3.57}) & 74.34 & 74.59 ({\color{c}+0.25})& 50.63 & 52.23 ({\color{c}+1.6}) \\
     \hline
     CIMMYT_3   & 60 & 47.56 & 47.84 ({\color{c}+0.28}) & 49.24 & 49.25 ({\color{c}+0.01}) & 35.01 & 39.12 ({\color{c}+4.11})\\
     \hline
     Ukyoto_1   & 60 & 48.91 & 54.55 ({\color{c}+5.64}) & 52.90 & 55.82 ({\color{c}+2.92}) & 39.31 & 42.62 ({\color{c}+3.31})\\
     \hline
     NAU_2   & 100 & 82.98 & 86.63 ({\color{c}+3.65}) & 88.80 & 88.64 ({\color{red}-0.16}) & 70.43 & 73.15 ({\color{c}+2.72}) \\
     \hline
     NAU_3   & 100 & 86.00 & 90.14 ({\color{c}+4.14}) & 91,41 & 91,51 ({\color{c}+0.10}) & 76.45 & 78.12 ({\color{c}+1.67}) \\
     \hline
     ARC_1   & 30 & 70.81 & 74.49 ({\color{c}+4.31}) & 70.55 & 72.14 ({\color{c}+1.59}) & 52.03 & 56.62 ({\color{c}+2.68})\\
     \hline\hline
     \multicolumn{2}{|c|}{\textbf{GWHD - All}} & \textbf{56.24} & \textbf{57.42} ({\color{c}\textbf{+1.18}})  & \textbf{59.24} & \textbf{60.67} ({\color{c}\textbf{+1.43}})& \textbf{44.44} & \textbf{46.54} ({\color{c}\textbf{+2.10}}) \\
     \hline
    \end{tabular}}
    \end{center}
    \label{tab:domain_specific_analysis}
\end{table*}

\textbf{Ablation studies.} Table \ref{tab:GWHD_SCB2} shows results from repeating the experiments from Tables \ref{tab:cross_dataset_generalisation} (selecting two out of four target domains) and \ref{tab:domain_specific_analysis}, while relying on a partial loss function to optimise the network weights. These show that best performance was achieved when all proposed loss functions were combined. It can be seen that while performing alignment across all domains simultaneously at both image and  instance level (second row) improves performance, as does domain-specific alignment (third row), combination of all proposed losses, which fully addresses both covariate and concept shift, is needed to achieve best performance.

\begin{table*}[ht]
    \caption{
    Ablation study on GHWD, \emph{IDD} (I), \emph{Cityscapes} (C) and \emph{BDD10K} (B). Left and right of $\longrightarrow$ indicate source and target datasets). BBA: Bounding Box Alignment. CCA: Class Conditional Alignment. Best results are bold. Results are reported in terms mAP values.} 
    \begin{center}
    \resizebox{\textwidth}{!}{
    \begin{tabular}{|c|c|c|c|c|c|c|c|c|c|c|c|}
    \hline
    Baseline&BBA&CCA& \multicolumn{3}{c|}{GWHD} & \multicolumn{3}{c|}{(A, B, C) $\rightarrow$ I} & \multicolumn{3}{c|}{(B, C, I) $\rightarrow$ A}\\
    \hline
    
     $L_{cls}$&$L_{dadv}$&$L_{erc}$& \multirow{3}{*}{\fcos} & \multirow{3}{*}{\yolo} & Faster & \multirow{3}{*}{\fcos} & \multirow{3}{*}{\yolo} & Faster & \multirow{3}{*}{\fcos} & \multirow{3}{*}{\yolo} & Faster \\
     $L_{reg}$&$L_{dinst}$&$L_{cel}$&  & & \multirow{2}{*}{R-CNN} & & & \multirow{2}{*}{R-CNN} & & & \multirow{2}{*}{R-CNN} \\
     &$L_{cst}$&&  & & & & & & & &  \\
     \hline

     +&-&-
     & 59.24 & 44.44 & 56.24 & 24.65 & 13.92 & 30.03 & 30.39 & 21.20 & 34.75 \\ 
    \hline    
    +&+&-
     & 59.45 & 45.62  & 56.80 & 24.73 & 13.99 & 30.10 & 30.42 & 21.52 & 35.12 \\ 
    \hline
    +&-&+ & 59.85 & 45.12 & 56.54 & 24.70 & 14.02 & 30.19 & 30.51 & 21.75 & 35.25 \\  
    \hline
    +&+&+ & \textbf{60.67} & \textbf{46.54} & \textbf{57.42} & \textbf{24.90} & \textbf{14.43} & \textbf{30.38} & \textbf{30.93} & \textbf{22.29} & \textbf{36.00} \\  
    \hline
    \end{tabular}}
    \end{center}
    \label{tab:GWHD_SCB2}
\end{table*}

\begin{figure}[t]
    \begin{center}
    \subfloat[]{\label{fig.dgfcos}\includegraphics[width=0.33\textwidth]{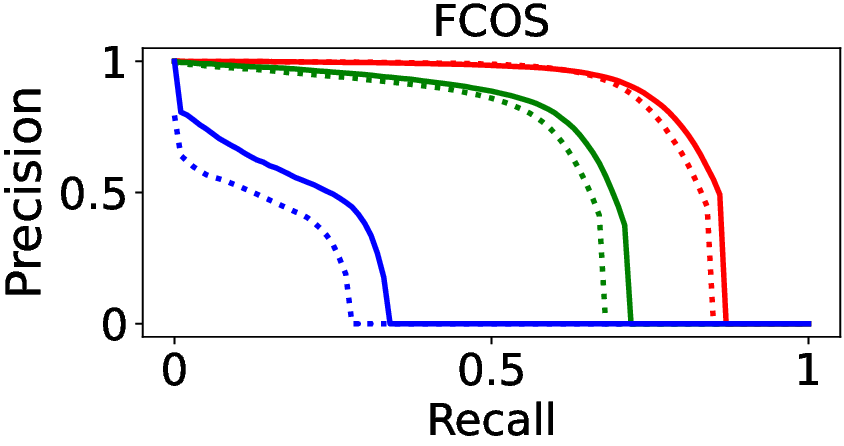}}
    \subfloat[]{\label{fig.dgyolo}\includegraphics[width=0.33\textwidth]{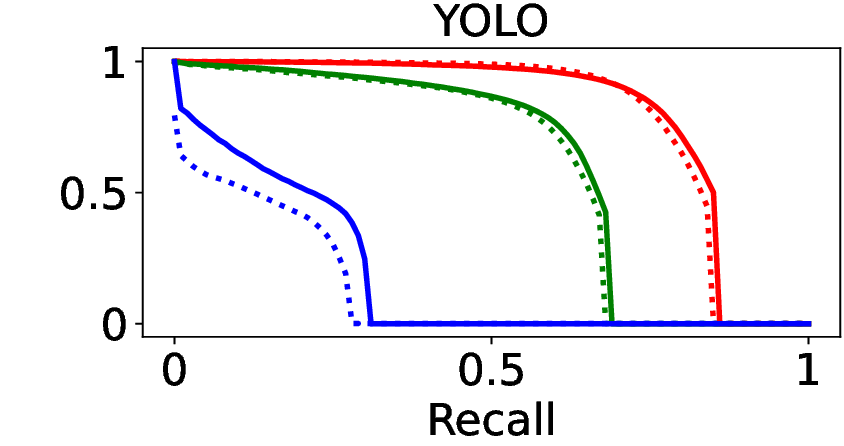}}
    \subfloat[]{\label{fig.dgfrcnn}\includegraphics[width=0.33\textwidth]{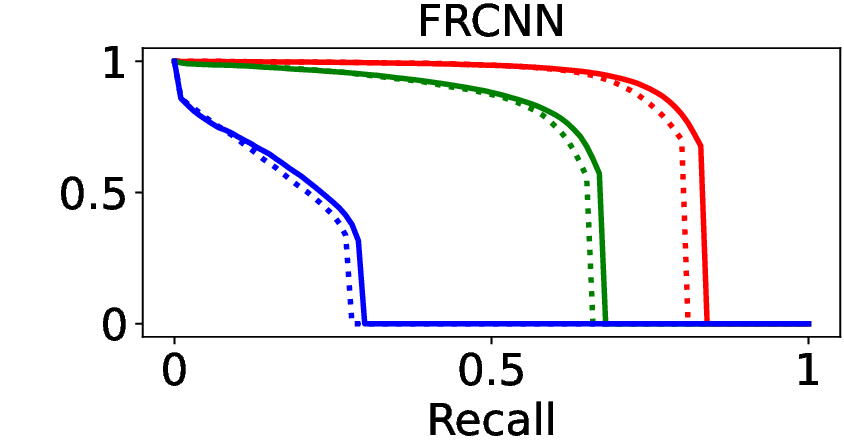}}
    \end{center}
    \caption{Precision-recall curves for GWHD. (a) \dgfcos vs \fcos, (b) \dgyolo vs \yolo, and (c) \dgfrcnn vs \frcnn. Color codes: IoU = 0.1 \crule[d1]{0.5cm}{0.25cm}, IoU = 0.5 \crule[c]{0.5cm}{0.25cm} and IoU = 0.75 \crule[d3]{0.5cm}{0.25cm}.}
    \label{fig:improved_localisation}
\end{figure}

To further study the effect of our proposed method on the three baseline architectures, we consider the performance of our models on the \emph{GWHD} dataset at different $\operatorname{IoU}$ levels. The single-class dataset was selected for this experiment, as in this case, $\operatorname{AP}_C$ is equivalent to the overall $\operatorname{mAP}$. We show the \emph{precision-recall (PR) curves} (which are the basis of calculating $\operatorname{AP}_C$) across three different IoU levels (10\%, 50\% and 75\%) for all models in Fig.\@~\ref{fig:improved_localisation}. The area under the curve on all PR curves at all $\operatorname{IoU}$ levels is higher when adding our proposed modules (full lines) to all baseline models (dotted lines), indicating that our approach consistently improves detector localisation.

\textbf{Comparison with existing approaches.} A comparison of \dgyolo with the approach proposed by \citet{liu2020towards} is shown in Table \ref{tab:URPC2019}. It can be seen that our proposed \dgyolo improves performance over the existing approach \cite{liu2016ssd} by $1.35 \operatorname{mAP}$, improving the performance on 5 out of 6 classes in the dataset. 
Table \ref{tab:comparison_against_lin_etal} shows the comparison of \dgfrcnn with the disentanglement-based approach \cite{lin2021domain}, where we report an improvement in $\operatorname{mAP}$ on 2 out of 3 datasets (\emph{BDD100k}: $+2.2$, \emph{FoggyCityscapes}: $+1.6$, \emph{Cityscapes}: $-1.3$). Individual class performance indicates that our proposed \dgfrcnn struggles on the minority class (\emph{train}), impacting the overall $\operatorname{mAP}$ on \emph{Cityscapes} and \emph{FoggyCityscapes}. 
We do not perform these experiments using any other baseline models, as the existing approaches \cite{liu2020towards,lin2021domain} only apply to a single baseline architecture (\yolo and \frcnn respectively).

\begin{table*}[ht]
    \caption{Quantitative analysis on  UPRC2019. Best results are bold. Scores show average performances for each class across 8  domains.
    } 
    \begin{center}
    \begin{tabular}{|c|c|c|c|c|c|c|}
    \hline
    Method & echinus & starfish & holothurian & scallop & waterweeds & mAP \\
    \hline
    baseline(\yolo) & 53.51 & 7.32 & 11.15 & 9.89 & 0.00 & 16.37 \\
    \hline
    WQT-only & 60.98 & 17.08 & 33.29 & 39.02 & 2.38 & 30.55 \\
    \hline
    \frcnn + FPN  & 29.49 & 5.91 & 9.13 & 1.07 & 10.40 & 11.23 \\
    \hline
    SSD512 & 26.62 & 14.44 & 18.07 & 1.41 & 14.5 & 15.22 \\
    \hline
    SSD300 & 27.31 & 14.57 & 13.62 & 3.01 & 2.98 & 12.31 \\
    \hline
    \cite{liu2020towards} & 63.84 & 27.37 & \textbf{35.64} & 36.88 & 5.11 & 33.77 \\
     \hline
    \dgyolo (Ours) & \textbf{65.21} & \textbf{28.10} & 25.10 & \textbf{42.10} & \textbf{15.10} & \textbf{35.12}   \\
    \hline
    \end{tabular}
    \end{center}
    \label{tab:URPC2019}
\end{table*}

\begin{table*}[ht]
\caption{Comparison of the proposed approach against that of Lin~\etal~\cite{lin2021domain}, using Cityscapes (C), Foggy Cityscapes (F), and BDD100k (B).}

\begin{center}
\small{
\begin{tabular}{|c|c|cccccccc|c|}
    \hline
    DG Setting & Methods & person & rider & car & truck & bus & train & motor & bike & mAP \\
    \hline
    \multirow{2}{*}{(F, B) $\rightarrow$ C} & \cite{lin2021domain} & 43.6 & 46.2 & 63.2 & \textbf{41.9} & \textbf{60.9} & \textbf{51.1} & 36.0 & 41.3 & \textbf{47.9}  \\
                                \cline{2-11}
                                & \dgfrcnn (Ours) & \textbf{51.7} &  \textbf{47.2} & \textbf{71.7} & 38.7 & 54.2 & 29.0 & \textbf{37.4} & \textbf{42.9} & 46.6    \\  
                                
                                \cline{2-11}
    \hline
    \multirow{2}{*}{(C, B) $\rightarrow$ F} & \cite{lin2021domain} & 31.8 & 38.4 & 49.3 & 27.7 & \textbf{35.7} & \textbf{26.5} & 24.8 & 33.1 & 33.4 \\
    \cline{2-11}
                                & \dgfrcnn (Ours) & \textbf{38.1} & \textbf{41.3} & \textbf{51.1} & \textbf{31.1} & 31.4 & 15.8 & \textbf{31.7} & \textbf{39.7} & \textbf{35.0}    \\ 
                
                                \cline{2-11}
    \hline
    \multirow{2}{*}{(F, C) $\rightarrow$ B} & \cite{lin2021domain} & 34.5 & 30.4 & 44.2 & \textbf{21.2} & \textbf{19.0} & 0.0 & 9.2 & \textbf{22.8} & 22.7  \\
    \cline{2-11}
                                & \dgfrcnn (Ours) & \textbf{44.8} & 30.4 & \textbf{59.4} & 10.8 & 14.4 & \textbf{0.1} & \textbf{22.9} & 16.7 & \textbf{24.9}  \\
                                
                                \cline{2-11}
    \hline
\end{tabular}}
\end{center}
\label{tab:comparison_against_lin_etal}
\end{table*}

\section{Conclusion}
\label{sec:conclusion_future_works}
We present a domain generalisation approach for object detection addresses both constituent components of domain shift (covariate and concept), based on a rigorous mathematical analysis. This is implemented as feature alignment achieved through GRL \cite{ganin2016domain}. 
The proposed method can be applied to both single-stage (\yolo, \fcos) and two-stage (\frcnn) object detection architectures, as the first architecture-agnostic DG-OD approach. Comprehensive evaluation on autonomous driving and precision agriculture datasets shows improved domain generalisation and localisation performance compared to baseline architectures. Ablations show that all proposed modules are required to address domain shift in full, however one notable drawback of our method is sensitivity to the regularisation constants $\alpha_3$ and $\alpha_4$. The method outperforms state-of-the-art approaches \cite{liu2020towards,lin2021domain} on their own benchmarks, and we propose a new and extended benchmark setup for evaluating domain generalisation approaches for object detection in autonomous driving, relying on 4 distinct datasets. 
Future work should focus on applying an automatic approach to determine the regularisation constants $\alpha$ \cite{liebel2018auxiliary}. Other possible directions include replacing GRL \cite{ganin2016domain} with different feature alignment techniques  \cite{li2018deep,ghifary2016scatter,mahajan2021domain,li2020domain,li2018domain} and using explicit constraints to eliminate domain specific bias.

\section*{Acknowledgements}
EA was supported by the Scientific and Technological Research Council of Türkiye (TUBITAK) under Grant Number 123R108. KS was supported by Lincoln Agri-Robotics in Expanding Excellence in England (E3).

\bibliographystyle{unsrtnat}  
\bibliography{references}  

\begin{thebibliography}{93}
\providecommand{\natexlab}[1]{#1}
\providecommand{\url}[1]{\texttt{#1}}
\expandafter\ifx\csname urlstyle\endcsname\relax
  \providecommand{\doi}[1]{doi: #1}\else
  \providecommand{\doi}{doi: \begingroup \urlstyle{rm}\Url}\fi

\bibitem[Jocher et~al.(2020)]{jocher2020yolov5}
G.~Jocher et~al.
\newblock {ultralytics/yolov5: v3.1 - Bug Fixes and Performance Improvements},
  October 2020.

\bibitem[Tan et~al.(2020)Tan, Pang, and Le]{tan2020efficientdet}
M.~Tan, R.~Pang, and Q.~V. Le.
\newblock Efficientdet: Scalable and efficient object detection.
\newblock In \emph{Proc.~of IEEE Conf. Comp. Vis. Patt. Recog.}, pages
  10781--10790, 2020.

\bibitem[Liu et~al.(2016)Liu, Anguelov, Erhan, Szegedy, Reed, Fu, and
  Berg]{liu2016ssd}
W.~Liu, D.~Anguelov, D.~Erhan, C.~Szegedy, S.~Reed, C-Y. Fu, and A.~C. Berg.
\newblock {SSD}: Single shot multibox detector.
\newblock In \emph{Proc.~Eur. Conf. Comput. Vis.}, pages 21--37, 2016.

\bibitem[Ren et~al.(2016)Ren, He, Girshick, and Sun]{ren2015faster}
S.~Ren, K.~He, R.~Girshick, and J.~Sun.
\newblock Faster {R-CNN}: Towards real-time object detection with region
  proposal networks.
\newblock \emph{IEEE Trans. Pattern Ana.~Mach.~Intel.}, 39\penalty0
  (6):\penalty0 1137--1149, 2016.

\bibitem[Lin et~al.(2017)Lin, Doll{\'a}r, Girshick, He, Hariharan, and
  Belongie]{lin2017feature}
T-Y. Lin, P.~Doll{\'a}r, R.~Girshick, K.~He, B.~Hariharan, and S.~Belongie.
\newblock Feature pyramid networks for object detection.
\newblock In \emph{Proc.~of IEEE Conf. Comp. Vis. Patt. Recog.}, pages
  2117--2125, 2017.

\bibitem[Dai et~al.(2016)Dai, Li, He, and Sun]{jifeng2016rfcn}
J.~Dai, Y.~Li, K.~He, and J.~Sun.
\newblock {R-FCN}: Object detection via region-based fully convolutional
  networks.
\newblock In D.~Lee, M.~Sugiyama, U.~Luxburg, I.~Guyon, and R.~Garnett,
  editors, \emph{Proc.~Adv. Neural Inform. Process. Syst.}, volume~29. Curran
  Associates, Inc., 2016.

\bibitem[Everingham et~al.(2010)Everingham, {Van Gool}, Williams, Winn, and
  Zisserman]{everingham2010pascal}
M.~Everingham, L.~{Van Gool}, C.~Williams, J.~Winn, and A.~Zisserman.
\newblock The {P}ascal visual object classes ({VOC}) challenge.
\newblock \emph{Int. J. Comput. Vis.}, 88\penalty0 (2):\penalty0 303--338,
  2010.

\bibitem[Lin et~al.(2014)]{lin2014microsoft}
T-Y. Lin et~al.
\newblock Microsoft {COCO}: Common objects in context.
\newblock In \emph{Proc.~Eur. Conf. Comput. Vis.}, pages 740--755, 2014.

\bibitem[Recht et~al.(2019)Recht, Roelofs, Schmidt, and
  Shankar]{recht2019imagenet}
B.~Recht, R.~Roelofs, L.~Schmidt, and V.~Shankar.
\newblock Do imagenet classifiers generalize to imagenet?
\newblock In \emph{Proc.~Int. Conf. Mach. Learn.}, pages 5389--5400, 2019.

\bibitem[Hendrycks and Dietterich(2019)]{hendrycks2019benchmarking}
D.~Hendrycks and T.~Dietterich.
\newblock Benchmarking neural network robustness to common corruptions and
  perturbations.
\newblock \emph{arXiv:1903.12261}, 2019.

\bibitem[Zhou et~al.(2021)Zhou, Liu, Qiao, Xiang, and Loy]{zhou2021domain}
K.~Zhou, Z.~Liu, Y.~Qiao, T.~Xiang, and C.~C. Loy.
\newblock Domain generalization: A survey.
\newblock \emph{arXiv:2103.02503}, 2021.

\bibitem[Koh et~al.(2021)]{koh2021wilds}
P.~W. Koh et~al.
\newblock Wilds: A benchmark of in-the-wild distribution shifts.
\newblock In \emph{Proc.~Int. Conf. Mach. Learn.}, pages 5637--5664, 2021.

\bibitem[{Moreno-Torres} et~al.(2012){Moreno-Torres}, Raeder,
  {Alaiz-Rodr{\'\i}guez}, Chawla, and Herrera]{moreno2012unifying}
J.~G. {Moreno-Torres}, T.~Raeder, R.~{Alaiz-Rodr{\'\i}guez}, N.~V. Chawla, and
  F.~Herrera.
\newblock A unifying view on dataset shift in classification.
\newblock \emph{Patt. Recogn.}, 45\penalty0 (1):\penalty0 521--530, 2012.

\bibitem[Liu et~al.(2020)Liu, Song, and Ding]{liu2020towards}
H.~Liu, P.~Song, and R.~Ding.
\newblock Towards domain generalization in underwater object detection.
\newblock In \emph{Proc. IEEE Int. Conf. Image Process.}, pages 1971--1975,
  2020.

\bibitem[Chen et~al.(2018)Chen, Li, Sakaridis, Dai, and {Van
  Gool}]{chen2018domain}
Y.~Chen, W.~Li, C.~Sakaridis, D.~Dai, and L.~{Van Gool}.
\newblock Domain adaptive {F}aster {R-CNN} for object detection in the wild.
\newblock In \emph{Proc.~of IEEE Conf. Comp. Vis. Patt. Recog.}, pages
  3339--3348, 2018.

\bibitem[Seemakurthy et~al.(2023{\natexlab{a}})Seemakurthy, Bosilj, Aptoula,
  and Fox]{seemakurthy2023icra}
K.~Seemakurthy, P.~Bosilj, E.~Aptoula, and C.~Fox.
\newblock Domain generalised fully convolutional one stage detection.
\newblock In \emph{Proc. IEEE Int. Conf. on Robotics and Autom.},
  2023{\natexlab{a}}.

\bibitem[Seemakurthy et~al.(2023{\natexlab{b}})Seemakurthy, Fox, Aptoula, and
  Bosilj]{seemakurthy2023aaai}
K.~Seemakurthy, C.~Fox, E.~Aptoula, and P.~Bosilj.
\newblock Domain generalised {Faster} {R-CNN}.
\newblock In \emph{Proc. of the AAAI Conf.}, 2023{\natexlab{b}}.

\bibitem[Lin et~al.(2021)Lin, Yuan, Zhao, Sun, Wang, and Cai]{lin2021domain}
C.~Lin, Z.~Yuan, S.~Zhao, P.~Sun, C.~Wang, and J.~Cai.
\newblock Domain-invariant disentangled network for generalizable object
  detection.
\newblock In \emph{Proc.~of IEEE Conf. Comp. Vis.}, pages 8771--8780, 2021.

\bibitem[Dalal and Triggs(2005)]{dalal2005histograms}
N.~Dalal and B.~Triggs.
\newblock Histograms of oriented gradients for human detection.
\newblock In \emph{Proc.~of IEEE Conf. Comp. Vis. Patt. Recog.}, volume~1,
  pages 886--893, 2005.

\bibitem[Felzenszwalb et~al.(2009)Felzenszwalb, Girshick, McAllester, and
  Ramanan]{felzenszwalb2009object}
P.~Felzenszwalb, R.~Girshick, D.~McAllester, and D.~Ramanan.
\newblock Object detection with discriminatively trained part-based models.
\newblock \emph{IEEE Trans. Pattern Ana.~Mach.~Intel.}, 32\penalty0
  (9):\penalty0 1627--1645, 2009.

\bibitem[Viola and Jones(2001)]{viola2001rapid}
P.~Viola and M.~Jones.
\newblock Rapid object detection using a boosted cascade of simple features.
\newblock In \emph{Proc.~of IEEE Conf. Comp. Vis. Patt. Recog.}, volume~1,
  2001.

\bibitem[LeCun et~al.(1998)LeCun, Bottou, Bengio, and
  Haffner]{lecun1998gradient}
Y.~LeCun, L.~Bottou, Y.~Bengio, and P.~Haffner.
\newblock Gradient-based learning applied to document recognition.
\newblock \emph{Proc. of the IEEE}, 86\penalty0 (11):\penalty0 2278--2324,
  1998.

\bibitem[Krizhevsky et~al.(2012)Krizhevsky, Sutskever, and
  Hinton]{krizhevsky2012imagenet}
A.~Krizhevsky, I.~Sutskever, and G.~E. Hinton.
\newblock Imagenet classification with deep convolutional neural networks.
\newblock \emph{Proc.~Adv. Neural Inform. Process. Syst.}, 25:\penalty0
  1097--1105, 2012.

\bibitem[Redmon and Farhadi(2018)]{redmon2018yolov3}
J.~Redmon and A.~Farhadi.
\newblock Yolov3: An incremental improvement.
\newblock \emph{arXiv:1804.02767}, 2018.

\bibitem[Redmon et~al.(2016)Redmon, Divvala, Girshick, and
  Farhadi]{redmon2016you}
J.~Redmon, S.~Divvala, R.~Girshick, and A.~Farhadi.
\newblock You only look once: Unified, real-time object detection.
\newblock In \emph{Proc.~of IEEE Conf. Comp. Vis. Patt. Recog.}, pages
  779--788, 2016.

\bibitem[Tian et~al.(2019)Tian, Shen, Chen, and He]{tian2019fcos}
Z.~Tian, C.~Shen, H.~Chen, and T.~He.
\newblock {FCOS}: Fully convolutional one-stage object detection.
\newblock In \emph{Proc.~of IEEE Conf. Comp. Vis.}, pages 9627--9636, 2019.

\bibitem[Carion et~al.(2020)Carion, Massa, Synnaeve, Usunier, Kirillov, and
  Zagoruyko]{carion2020end}
N.~Carion, F.~Massa, G.~Synnaeve, N.~Usunier, A.~Kirillov, and S.~Zagoruyko.
\newblock End-to-end object detection with transformers.
\newblock In \emph{Proc.~Eur. Conf. Comput. Vis.}, pages 213--229, 2020.

\bibitem[He and Todorovic(2022)]{He2022CVPR}
L.~He and S.~Todorovic.
\newblock {DESTR}: Object detection with split transformer.
\newblock In \emph{Proc.~of IEEE Conf. Comp. Vis. Patt. Recog.}, 2022.

\bibitem[Zhang et~al.(2023)Zhang, Luo, Tian, Zhang, Zhang, and Lu]{10204463}
G.~Zhang, Z.~Luo, Z.~Tian, J.~Zhang, X.~Zhang, and S.~Lu.
\newblock Towards efficient use of multi-scale features in transformer-based
  object detectors.
\newblock In \emph{Proc.~of IEEE Conf. Comp. Vis. Patt. Recog.}, pages
  6206--6216, 2023.

\bibitem[Huang et~al.(2023)Huang, Dai, Xiang, Wang, Chen, Qin, and
  Xiong]{10204055}
Z.~Huang, H.~Dai, T-Z. Xiang, S.~Wang, H-X. Chen, J.~Qin, and H.~Xiong.
\newblock Feature shrinkage pyramid for camouflaged object detection with
  transformers.
\newblock In \emph{Proc.~of IEEE Conf. Comp. Vis. Patt. Recog.}, pages
  5557--5566, 2023.

\bibitem[Girshick et~al.(2014)Girshick, Donahue, Darrell, and
  Malik]{girshick2014rich}
R.~Girshick, J.~Donahue, T.~Darrell, and J.~Malik.
\newblock Rich feature hierarchies for accurate object detection and semantic
  segmentation.
\newblock In \emph{Proc.~of IEEE Conf. Comp. Vis. Patt. Recog.}, pages
  580--587, 2014.

\bibitem[Girshick(2015)]{girshick2015fast}
R.~Girshick.
\newblock Fast {R-CNN}.
\newblock In \emph{Proc.~of IEEE Conf. Comp. Vis.}, pages 1440--1448, 2015.

\bibitem[Pan and Yang(2010)]{pan2010survey}
S.~J. Pan and Q.~Yang.
\newblock A survey on transfer learning.
\newblock \emph{IEEE Tran. on know. and data eng.}, 22\penalty0 (10):\penalty0
  1345--1359, 2010.

\bibitem[Deng et~al.(2009)Deng, Dong, Socher, Li, Li, and
  Fei-Fei]{deng2009imagenet}
J.~Deng, W.~Dong, R.~Socher, L-J. Li, K.~Li, and L.~Fei-Fei.
\newblock Imagenet: A large-scale hierarchical image database.
\newblock In \emph{Proc.~of IEEE Conf. Comp. Vis. Patt. Recog.}, pages
  248--255, 2009.

\bibitem[Zhuang et~al.(2020)]{zhuang2020comprehensive}
F.~Zhuang et~al.
\newblock A comprehensive survey on transfer learning.
\newblock \emph{Proceedings of the IEEE}, 109\penalty0 (1):\penalty0 43--76,
  2020.

\bibitem[Bosilj et~al.(2020)Bosilj, Aptoula, Duckett, and
  Cielniak]{bosilj2020transfer}
P.~Bosilj, E.~Aptoula, T.~Duckett, and G.~Cielniak.
\newblock Transfer learning between crop types for semantic segmentation of
  crops versus weeds in precision agriculture.
\newblock \emph{J. Field Robotics}, 37\penalty0 (1):\penalty0 7--19, 2020.

\bibitem[Ghazi et~al.(2017)Ghazi, Yanikoglu, and Aptoula]{ghazi2017plant}
M.~M. Ghazi, B.~Yanikoglu, and E.~Aptoula.
\newblock Plant identification using deep neural networks via optimization of
  transfer learning parameters.
\newblock \emph{Neurocomputing}, 235:\penalty0 228--235, 2017.

\bibitem[Niu et~al.(2020)Niu, Liu, Wang, and Song]{niu2020decade}
S.~Niu, Y.~Liu, J.~Wang, and H.~Song.
\newblock A decade survey of transfer learning (2010--2020).
\newblock \emph{IEEE Trans.~on Artific.~Intell.}, 1\penalty0 (2):\penalty0
  151--166, 2020.

\bibitem[Wang et~al.(2019)Wang, Liew, Zou, Zhou, and Feng]{wang2019panet}
K.~Wang, J.~H. Liew, Y.~Zou, D.~Zhou, and J.~Feng.
\newblock Panet: Few-shot image semantic segmentation with prototype alignment.
\newblock In \emph{Proc.~of IEEE Conf. Comp. Vis.}, pages 9197--9206, 2019.

\bibitem[Jing and Tian(2020)]{jing2020self}
L.~Jing and Y.~Tian.
\newblock Self-supervised visual feature learning with deep neural networks: A
  survey.
\newblock \emph{IEEE Trans. Pattern Ana.~Mach.~Intel.}, 43\penalty0
  (11):\penalty0 4037--4058, 2020.

\bibitem[Noroozi and Favaro(2016)]{noroozi2016unsupervised}
M.~Noroozi and P.~Favaro.
\newblock Unsupervised learning of visual representations by solving jigsaw
  puzzles.
\newblock In \emph{Proc.~Eur. Conf. Comput. Vis.}, pages 69--84, 2016.

\bibitem[Hindel et~al.(2023)Hindel, Gosala, Bregler, and
  Valada]{hindel2023inod}
J.~Hindel, N.~Gosala, K.~Bregler, and A.~Valada.
\newblock {INoD}: Injected noise discriminator for self-supervised
  representation learning in agricultural fields.
\newblock \emph{arXiv:2303.18101}, 2023.

\bibitem[Ilteralp et~al.(2021)Ilteralp, Ariman, and Aptoula]{ilteralp2021deep}
M.~Ilteralp, S.~Ariman, and E.~Aptoula.
\newblock A deep multitask semisupervised learning approach for chlorophyll-a
  retrieval from remote sensing images.
\newblock \emph{Remote Sensing}, 14\penalty0 (1):\penalty0 18, 2021.

\bibitem[Ullah et~al.(2023)Ullah, Usman, Latif, Khan, and Gwak]{ullah2023ssmd}
Z.~Ullah, M.~Usman, S.~Latif, A.~Khan, and J.~Gwak.
\newblock {SSMD-UNet}: semi-supervised multi-task decoders network for diabetic
  retinopathy segmentation.
\newblock \emph{Scientific Reports}, 13\penalty0 (1):\penalty0 9087, 2023.

\bibitem[Li et~al.(2022)Li, Liu, and Bilen]{li2022learning}
W-H. Li, X.~Liu, and H.~Bilen.
\newblock Learning multiple dense prediction tasks from partially annotated
  data.
\newblock In \emph{Proc.~of IEEE Conf. Comp. Vis. Patt. Recog.}, pages
  18879--18889, 2022.

\bibitem[Yang and Hospedales(2016)]{yang2016deep}
Y.~Yang and T.~M. Hospedales.
\newblock Deep multi-task representation learning: A tensor factorisation
  approach.
\newblock In \emph{Int. Conf. Learn. Represent.}, 2016.

\bibitem[Wang et~al.(2020{\natexlab{a}})Wang, Yu, Li, Fu, and
  Heng]{wang2020learning}
S.~Wang, L.~Yu, C.~Li, C-W. Fu, and P-A. Heng.
\newblock Learning from extrinsic and intrinsic supervisions for domain
  generalization.
\newblock In \emph{Proc.~Eur. Conf. Comput. Vis.}, pages 159--176,
  2020{\natexlab{a}}.

\bibitem[Tobin et~al.(2017)Tobin, Fong, Ray, Schneider, Zaremba, and
  Abbeel]{tobin2017domain}
J.~Tobin, R.~Fong, A.~Ray, J.~Schneider, W.~Zaremba, and P.~Abbeel.
\newblock Domain randomization for transferring deep neural networks from
  simulation to the real world.
\newblock In \emph{Int. Conf. on Intell. Robots and Syst,}, pages 23--30, 2017.

\bibitem[Yue et~al.(2019)Yue, Zhang, Zhao, Sangiovanni-Vincentelli, Keutzer,
  and Gong]{yue2019domain}
X.~Yue, Y.~Zhang, S.~Zhao, A.~Sangiovanni-Vincentelli, K.~Keutzer, and B.~Gong.
\newblock Domain randomization and pyramid consistency: Simulation-to-real
  generalization without accessing target domain data.
\newblock In \emph{Proc.~of IEEE Conf. Comp. Vis.}, pages 2100--2110, 2019.

\bibitem[Saenko et~al.(2010)Saenko, Kulis, Fritz, and
  Darrell]{saenko2010adapting}
K.~Saenko, B.~Kulis, M.~Fritz, and T.~Darrell.
\newblock Adapting visual category models to new domains.
\newblock In \emph{Proc.~Eur. Conf. Comput. Vis.}, pages 213--226, 2010.

\bibitem[Blanchard et~al.(2011)Blanchard, Lee, and
  Scott]{blanchard2011generalizing}
G.~Blanchard, G.~Lee, and C.~Scott.
\newblock Generalizing from several related classification tasks to a new
  unlabeled sample.
\newblock \emph{Proc.~Adv. Neural Inform. Process. Syst.}, 24:\penalty0
  2178--2186, 2011.

\bibitem[Muandet et~al.(2013)Muandet, Balduzzi, and
  Sch{\"o}lkopf]{muandet2013domain}
K.~Muandet, D.~Balduzzi, and B.~Sch{\"o}lkopf.
\newblock Domain generalization via invariant feature representation.
\newblock In \emph{Proc.~Int. Conf. Mach. Learn.}, pages 10--18, 2013.

\bibitem[Khosla et~al.(2012)Khosla, Zhou, Malisiewicz, Efros, and
  Torralba]{khosla2012undoing}
A.~Khosla, T.~Zhou, T.~Malisiewicz, A.~A. Efros, and A.~Torralba.
\newblock Undoing the damage of dataset bias.
\newblock In \emph{Proc.~Eur. Conf. Comput. Vis.}, pages 158--171, 2012.

\bibitem[Xu et~al.(2014)Xu, Li, Niu, and Xu]{xu2014exploiting}
Z.~Xu, W.~Li, L.~Niu, and D.~Xu.
\newblock Exploiting low-rank structure from latent domains for domain
  generalization.
\newblock In \emph{Proc.~Eur. Conf. Comput. Vis.}, pages 628--643, 2014.

\bibitem[Fang et~al.(2013)Fang, Xu, and Rockmore]{fang2013unbiased}
C.~Fang, Y.~Xu, and D.~N. Rockmore.
\newblock Unbiased metric learning: On the utilization of multiple datasets and
  web images for softening bias.
\newblock In \emph{Proc.~of IEEE Conf. Comp. Vis.}, pages 1657--1664, 2013.

\bibitem[Ghifary et~al.(2015)Ghifary, Kleijn, Zhang, and
  Balduzzi]{ghifary2015domain}
M.~Ghifary, W.~B. Kleijn, M.~Zhang, and D.~Balduzzi.
\newblock Domain generalization for object recognition with multi-task
  autoencoders.
\newblock In \emph{Proc.~of IEEE Conf. Comp. Vis.}, pages 2551--2559, 2015.

\bibitem[Li et~al.(2017)Li, Yang, Song, and Hospedales]{li2017deeper}
D.~Li, Y.~Yang, Y-Z. Song, and T.~M. Hospedales.
\newblock Deeper, broader and artier domain generalization.
\newblock In \emph{Proc.~of IEEE Conf. Comp. Vis.}, pages 5542--5550, 2017.

\bibitem[Li et~al.(2018{\natexlab{a}})Li, Yang, Song, and
  Hospedales]{li2018learning}
D.~Li, Y.~Yang, Y-Z. Song, and T.~M. Hospedales.
\newblock Learning to generalize: Meta-learning for domain generalization.
\newblock In \emph{Proc. of the AAAI Conf.}, 2018{\natexlab{a}}.

\bibitem[Shankar et~al.(2018)Shankar, Piratla, Chakrabarti, Chaudhuri, Jyothi,
  and Sarawagi]{shankar2018generalizing}
S.~Shankar, V.~Piratla, S.~Chakrabarti, S.~Chaudhuri, P.~Jyothi, and
  S.~Sarawagi.
\newblock Generalizing across domains via cross-gradient training.
\newblock In \emph{Int. Conf. Learn. Represent.}, 2018.

\bibitem[Li et~al.(2019)Li, Zhang, Yang, Liu, Song, and
  Hospedales]{li2019episodic}
D.~Li, J.~Zhang, Y.~Yang, C.~Liu, Y-Z. Song, and T.~M. Hospedales.
\newblock Episodic training for domain generalization.
\newblock In \emph{Proc.~of IEEE Conf. Comp. Vis.}, pages 1446--1455, 2019.

\bibitem[Ganin et~al.(2016)]{ganin2016domain}
Y.~Ganin et~al.
\newblock Domain-adversarial training of neural networks.
\newblock \emph{J. Machine Learning Research}, 17\penalty0 (1):\penalty0
  2096--2030, 2016.

\bibitem[Li et~al.(2018{\natexlab{b}})]{li2018deep}
Y.~Li et~al.
\newblock Deep domain generalization via conditional invariant adversarial
  networks.
\newblock In \emph{ECCV}, pages 624--639, 2018{\natexlab{b}}.

\bibitem[Ghifary et~al.(2016)Ghifary, Balduzzi, Kleijn, and
  Zhang]{ghifary2016scatter}
M.~Ghifary, D.~Balduzzi, W.~B. Kleijn, and M.~Zhang.
\newblock Scatter component analysis: A unified framework for domain adaptation
  and domain generalization.
\newblock \emph{IEEE Trans. Pattern Ana.~Mach.~Intel.}, 39\penalty0
  (7):\penalty0 1414--1430, 2016.

\bibitem[Mahajan et~al.(2021)Mahajan, Tople, and Sharma]{mahajan2021domain}
D.~Mahajan, S.~Tople, and A.~Sharma.
\newblock Domain generalization using causal matching.
\newblock In \emph{Proc.~Int. Conf. Mach. Learn.}, pages 7313--7324, 2021.

\bibitem[Li et~al.(2020)Li, Wang, Wan, Wang, Li, and Kot]{li2020domain}
H.~Li, Y.~Wang, R.~Wan, S.~Wang, T-Q. Li, and A.~Kot.
\newblock Domain generalization for medical imaging classification with
  linear-dependency regularization.
\newblock \emph{Proc.~Adv. Neural Inform. Process. Syst.}, 33:\penalty0
  3118--3129, 2020.

\bibitem[Li et~al.(2018{\natexlab{c}})Li, Pan, Wang, and Kot]{li2018domain}
H.~Li, S.~J. Pan, S.~Wang, and A.~C. Kot.
\newblock Domain generalization with adversarial feature learning.
\newblock In \emph{Proc.~of IEEE Conf. Comp. Vis. Patt. Recog.}, pages
  5400--5409, 2018{\natexlab{c}}.

\bibitem[Ilse et~al.(2020)Ilse, Tomczak, Louizos, and Welling]{ilse2020diva}
M.~Ilse, J.~M. Tomczak, C.~Louizos, and M.~Welling.
\newblock Diva: Domain invariant variational autoencoders.
\newblock In \emph{Proc.~Med. Imag. with Deep Learn.}, pages 322--348, 2020.

\bibitem[Finn et~al.(2017)Finn, Abbeel, and Levine]{finn2017model}
C.~Finn, P.~Abbeel, and S.~Levine.
\newblock Model-agnostic meta-learning for fast adaptation of deep networks.
\newblock In \emph{Proc.~Int. Conf. Mach. Learn.}, pages 1126--1135, 2017.

\bibitem[Dou et~al.(2019)Dou, {Coelho de Castro}, Kamnitsas, and
  Glocker]{dou2019domain}
Q.~Dou, D.~{Coelho de Castro}, K.~Kamnitsas, and B.~Glocker.
\newblock Domain generalization via model-agnostic learning of semantic
  features.
\newblock \emph{Proc.~Adv. Neural Inform. Process. Syst.}, 32, 2019.

\bibitem[Du et~al.(2020)Du, Zhen, Shao, and Snoek]{du2020metanorm}
Y.~Du, X.~Zhen, L.~Shao, and C.~Snoek.
\newblock Metanorm: Learning to normalize few-shot batches across domains.
\newblock In \emph{Int. Conf. Learn. Represent.}, 2020.

\bibitem[Yuge et~al.(2022)]{shi2022gradient}
S.~Yuge et~al.
\newblock Gradient matching for domain generalization.
\newblock In \emph{Int. Conf. Learn. Represent.}, 2022.

\bibitem[Zhou et~al.(2020)Zhou, Yang, Hospedales, and Xiang]{zhou2020learning}
K.~Zhou, Y.~Yang, T.~M. Hospedales, and T.~Xiang.
\newblock Learning to generate novel domains for domain generalization.
\newblock In \emph{Proc.~Eur. Conf. Comput. Vis.}, pages 561--578, 2020.

\bibitem[Song et~al.(2019)Song, Yang, Song, Xiang, and
  Hospedales]{song2019generalizable}
J.~Song, Y.~Yang, Y-Z. Song, T.~Xiang, and T.~M. Hospedales.
\newblock Generalizable person re-identification by domain-invariant mapping
  network.
\newblock In \emph{Proc.~of IEEE Conf. Comp. Vis. Patt. Recog.}, pages
  719--728, 2019.

\bibitem[Wang et~al.(2020{\natexlab{b}})Wang, Han, Shan, and
  Chen]{wang2020cross}
G.~Wang, H.~Han, S.~Shan, and X.~Chen.
\newblock Cross-domain face presentation attack detection via multi-domain
  disentangled representation learning.
\newblock In \emph{Proc.~of IEEE Conf. Comp. Vis. Patt. Recog.}, pages
  6678--6687, 2020{\natexlab{b}}.

\bibitem[Jia et~al.(2020)Jia, Zhang, Shan, and Chen]{jia2020single}
Y.~Jia, J.~Zhang, S.~Shan, and X.~Chen.
\newblock Single-side domain generalization for face anti-spoofing.
\newblock In \emph{Proc.~of IEEE Conf. Comp. Vis. Patt. Recog.}, pages
  8484--8493, 2020.

\bibitem[Zhao et~al.(2020)Zhao, Gong, Liu, Fu, and Tao]{zhao2020domain}
S.~Zhao, M.~Gong, T.~Liu, H.~Fu, and D.~Tao.
\newblock Domain generalization via entropy regularization.
\newblock \emph{Proc.~Adv. Neural Inform. Process. Syst.}, 33, 2020.

\bibitem[Sch{\"o}lkopf et~al.(2012)Sch{\"o}lkopf, Janzing, Peters, Sgouritsa,
  Zhang, and Mooij]{scholkopf2012causal}
B.~Sch{\"o}lkopf, D.~Janzing, J.~Peters, E.~Sgouritsa, K.~Zhang, and J.~Mooij.
\newblock On causal and anticausal learning.
\newblock \emph{arXiv:1206.6471}, 2012.

\bibitem[Janzing and Sch{\"o}lkopf(2010)]{janzing2010causal}
D.~Janzing and B.~Sch{\"o}lkopf.
\newblock Causal inference using the algorithmic {Markov} condition.
\newblock \emph{IEEE Trans. Inf. Theory}, 56\penalty0 (10):\penalty0
  5168--5194, 2010.

\bibitem[Hu et~al.(2020)Hu, Zhang, Chen, and Chan]{hu2020domain}
S.~Hu, K.~Zhang, Z.~Chen, and L.~Chan.
\newblock Domain generalization via multidomain discriminant analysis.
\newblock In \emph{Uncertainty in Artificial Intelligence}, pages 292--302,
  2020.

\bibitem[Chen et~al.(2023)Chen, Song, Liu, Dai, Zhang, Ding, and
  Li]{chen2023achieving}
Y.~Chen, P.~Song, H.~Liu, L.~Dai, X.~Zhang, R.~Ding, and S.~Li.
\newblock Achieving domain generalization for underwater object detection by
  domain mixup and contrastive learning.
\newblock \emph{Neurocomput.}, 528\penalty0 (C):\penalty0 20--34, 2023.

\bibitem[Arjovsky et~al.(2019)Arjovsky, Bottou, Gulrajani, and
  {Lopez-Paz}]{arjovsky2019invariant}
M.~Arjovsky, L.~Bottou, I.~Gulrajani, and D.~{Lopez-Paz}.
\newblock Invariant risk minimization.
\newblock \emph{arXiv:1907.02893}, 2019.

\bibitem[Matsuura and Harada(2020)]{matsuura2020domain}
T.~Matsuura and T.~Harada.
\newblock Domain generalization using a mixture of multiple latent domains.
\newblock In \emph{Proc. of the AAAI Conf.}, volume~34, pages 11749--11756,
  2020.

\bibitem[Goodfellow et~al.(2014)Goodfellow, Pouget-Abadie, Mirza, Xu,
  Warde-Farley, Ozair, Courville, and Bengio]{goodfellow2014generative}
I.~J. Goodfellow, J.~Pouget-Abadie, M.~Mirza, B.~Xu, D.~Warde-Farley, S.~Ozair,
  A.~Courville, and Y.~Bengio.
\newblock Generative adversarial networks.
\newblock \emph{arXiv:1406.2661}, 2014.

\bibitem[Endres and Schindelin(2003)]{endres2003new}
D.~Endres and J.~Schindelin.
\newblock A new metric for probability distributions.
\newblock \emph{IEEE Trans. Inf. Theory}, 49\penalty0 (7):\penalty0 1858--1860,
  2003.

\bibitem[Gong et~al.(2019)Gong, Xu, Li, Zhang, and Batmanghelich]{gong2019twin}
M.~Gong, Y.~Xu, C.~Li, K.~Zhang, and K.~Batmanghelich.
\newblock Twin auxiliary classifiers {GAN}.
\newblock \emph{Proc.~Adv. Neural Inform. Process. Syst.}, 32:\penalty0 1328,
  2019.

\bibitem[Cordts et~al.(2016)]{cordts2016cityscapes}
M.~Cordts et~al.
\newblock The {C}ityscapes dataset for semantic urban scene understanding.
\newblock In \emph{Proc.~of IEEE Conf. Comp. Vis. Patt. Recog.}, pages
  3213--3223, 2016.

\bibitem[Sakaridis et~al.(2018)Sakaridis, Dai, and {Van Gool}]{foggy}
C.~Sakaridis, D.~Dai, and L.~{Van Gool}.
\newblock Semantic foggy scene understanding with synthetic data.
\newblock \emph{Int. J. Comput. Vis.}, 126\penalty0 (9):\penalty0 973--992, Sep
  2018.

\bibitem[Yu et~al.(2020)Yu, Chen, Wang, Xian, Chen, Liu, Madhavan, and
  Darrell]{bdd100k}
F.~Yu, H.~Chen, X.~Wang, W.~Xian, Y.~Chen, F.~Liu, V.~Madhavan, and T.~Darrell.
\newblock Bdd100k: A diverse driving dataset for heterogeneous multitask
  learning.
\newblock In \emph{Proc.~of IEEE Conf. Comp. Vis. Patt. Recog.}, pages
  2636--2645, 2020.

\bibitem[Sakaridis et~al.(2021)Sakaridis, Dai, and {Van Gool}]{acdc-dataset}
C.~Sakaridis, D.~Dai, and L.~{Van Gool}.
\newblock {ACDC}: The adverse conditions dataset with correspondences for
  semantic driving scene understanding.
\newblock In \emph{Proc.~of IEEE Conf. Comp. Vis.}, October 2021.

\bibitem[Varma et~al.(2019)Varma, Subramanian, Namboodiri, Chandraker, and
  Jawahar]{idd-dataset}
G.~Varma, A.~Subramanian, A.~Namboodiri, M.~Chandraker, and C.~V. Jawahar.
\newblock {IDD}: A dataset for exploring problems of autonomous navigation in
  unconstrained environments.
\newblock In \emph{IEEE Winter Conf. App. Comp. Vis.}, pages 1743--1751, 2019.

\bibitem[David et~al.(2021)]{david2021global}
E.~David et~al.
\newblock Global wheat head detection 2021: an improved dataset for
  benchmarking wheat head detection methods.
\newblock \emph{Plant Phenomics}, 2021, 2021.

\bibitem[David(2021)]{david2021dataset}
E.~David.
\newblock Global wheat head dataset 2021, July 2021.
\newblock URL \url{https://doi.org/10.5281/zenodo.5092309}.

\bibitem[Liebel and K{\"o}rner(2018)]{liebel2018auxiliary}
L.~Liebel and M.~K{\"o}rner.
\newblock Auxiliary tasks in multi-task learning.
\newblock \emph{arXiv preprint arXiv:1805.06334}, 2018.

\end{thebibliography}

\appendix

\end{document}